\documentclass{article}

\usepackage{tikz}
\usepackage[hyphens]{url}

\usepackage{microtype}
\usepackage{todonotes}
\usepackage{graphicx}
\usepackage{subfigure, subcaption}
\usepackage{booktabs, xcolor, amsmath, amssymb, multirow, makecell, array, placeins} 
\usepackage{enumitem}
\usepackage{hyperref}
\usepackage{xspace}

\newcommand{\framework}{BOOST\xspace}
\newcommand{\vanilla}{\emph{Vanilla-TP}\xspace}
\newcommand{\fullrank}{\emph{FullRank-TP}\xspace}

\DeclareMathAlphabet\mathbfcal{OMS}{cmsy}{b}{n}

\newcommand{\mat}[1]{\mathbf{#1}}
\newcolumntype{L}[1]{>{\raggedright\arraybackslash}p{#1}}

\usepackage[accepted]{mlsys2025}

\mlsystitlerunning{BOOST: BOttleneck-Optimized Scalable Training Framework for Low-Rank Large Language Models}

\AddToShipoutPictureBG*{
\AtPageUpperLeft{
  \hspace*{\paperwidth}
  \raisebox{-56pt}{
    \llap{
      \href{https://www.acm.org/publications/policies/artifact-review-and-badging-current}{
        \includegraphics[height=48pt]{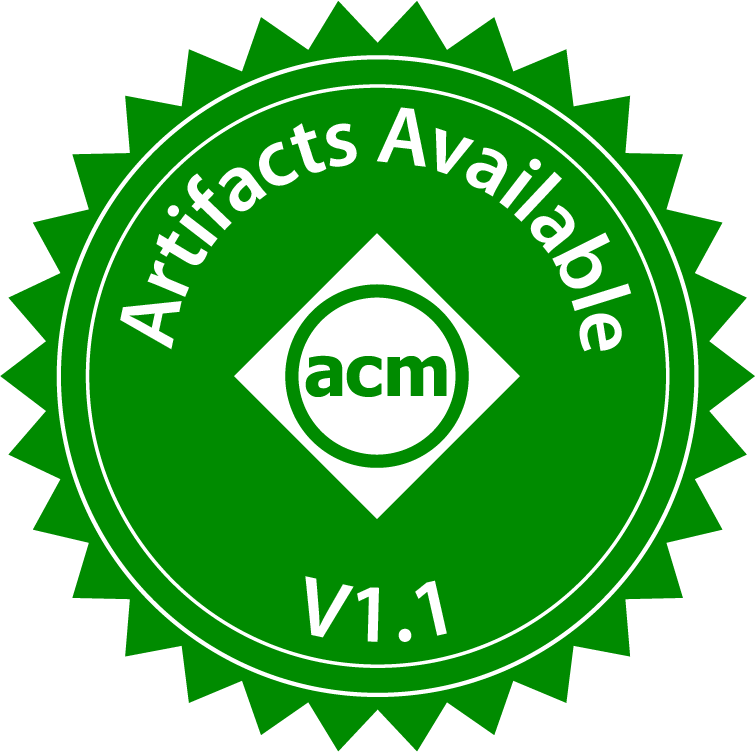}}
      \hspace{1pt}
      \href{https://www.acm.org/publications/policies/artifact-review-and-badging-current}{
        \includegraphics[height=48pt]{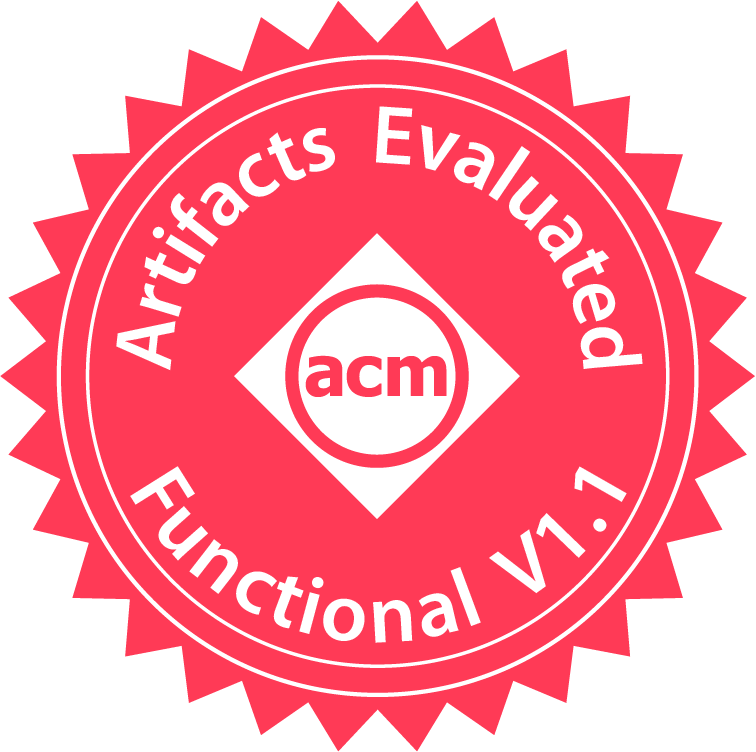}}
      \hspace{80pt}
    }
  }
}
}

\begin{document}

\twocolumn[
\mlsystitle{BOOST: BOttleneck-Optimized Scalable Training Framework for Low-Rank Large Language Models}



\mlsyssetsymbol{equal}{*}

\begin{mlsysauthorlist}
\mlsysauthor{Zhengyang Wang}{equal,ucsb}
\mlsysauthor{Ziyue Liu}{equal,ucsb}
\mlsysauthor{Ruijie Zhang}{ucsb}
\mlsysauthor{Avinash Maurya}{argonne}
\mlsysauthor{Paul Hovland}{argonne}
\mlsysauthor{Bogdan Nicolae}{argonne}
\mlsysauthor{Franck Cappello}{argonne}
\mlsysauthor{Zheng Zhang}{ucsb}
\end{mlsysauthorlist}

\mlsysaffiliation{ucsb}{University of California, Santa Barbara}
\mlsysaffiliation{argonne}{Argonne National Laboratory}

\mlsyscorrespondingauthor{Zheng Zhang}{zhengzhang@ece.ucsb.edu}

\mlsyskeywords{Pre-training, Distributed Training, 3D Parallelism, Model Compression, Machine Learning, MLSys}

\vskip 0.3in

\begin{abstract}
The scale of transformer model pre-training is constrained by the increasing computation and communication cost. Low-rank bottleneck architectures offer a promising solution to significantly reduce the training time and memory footprint with minimum impact on accuracy. Despite algorithmic efficiency, bottleneck architectures scale poorly under standard tensor parallelism. Simply applying 3D parallelism designed for full-rank
methods leads to excessive communication and poor GPU utilization. To address this limitation, we propose \framework{}, an efficient training framework tailored for large-scale low-rank bottleneck architectures. BOOST introduces a novel Bottleneck-aware Tensor Parallelism, and combines optimizations such as online-RMSNorm, linear layer grouping, and low-rank activation checkpointing to achieve end-to-end training speedup. Evaluations on different low-rank bottleneck architectures demonstrate that \framework{} achieves 1.46–1.91$\times$ speedup over full-rank model baselines and 1.87–2.27$\times$ speedup over low-rank model with naively integrated 3D parallelism, with improved GPU utilization and reduced communication overhead. Code available \href{https://github.com/Arcana-2236/BOOST}{here}.
\end{abstract}
]



\printAffiliationsAndNotice{\mlsysEqualContribution} 

\begin{figure*}[t]
  \centering
  \begin{minipage}{0.27\textwidth}
    \centering
    \includegraphics[width=\linewidth]{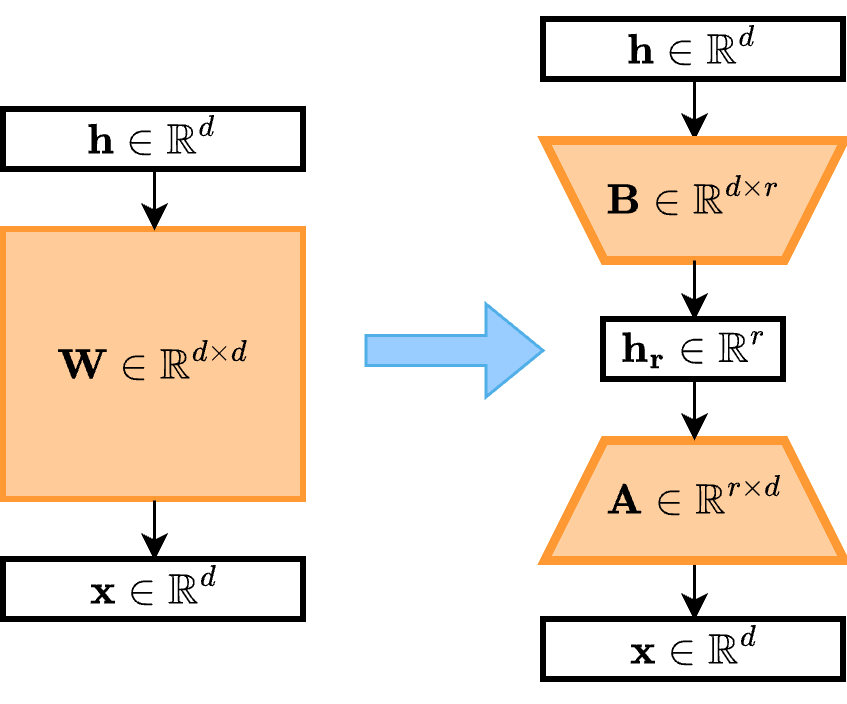}\\
  \end{minipage}\hfill
  \begin{minipage}{0.35\textwidth}
    \centering
    \includegraphics[width=\linewidth]{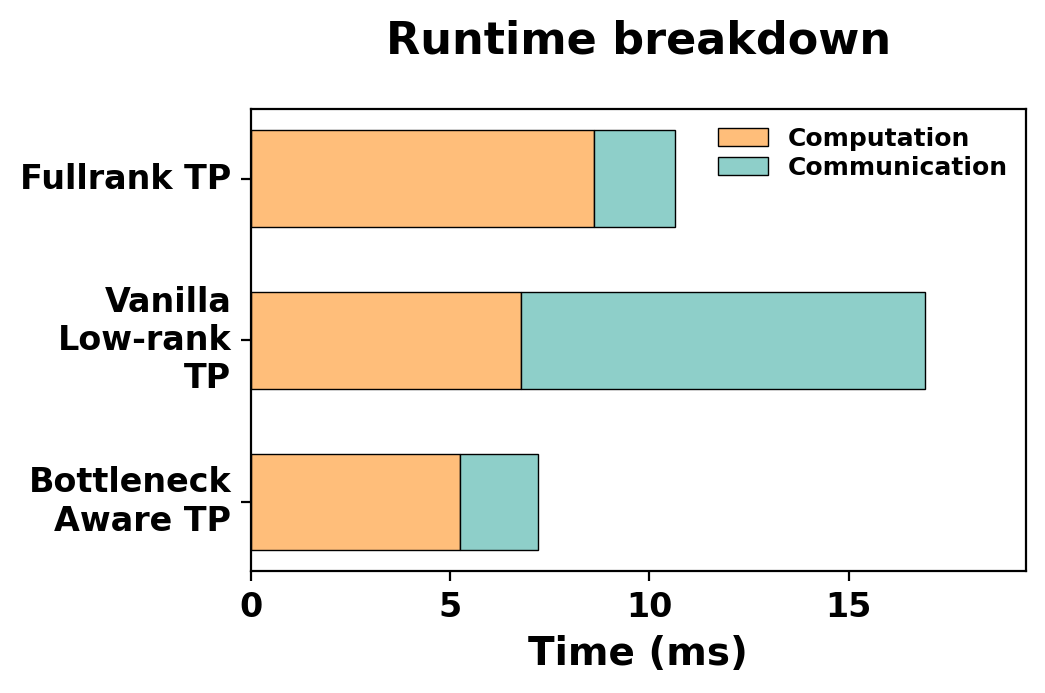}\\
  \end{minipage}\hfill
  \begin{minipage}{0.32\textwidth}
    \centering
    \includegraphics[width=\linewidth, height=0.8\linewidth, keepaspectratio]{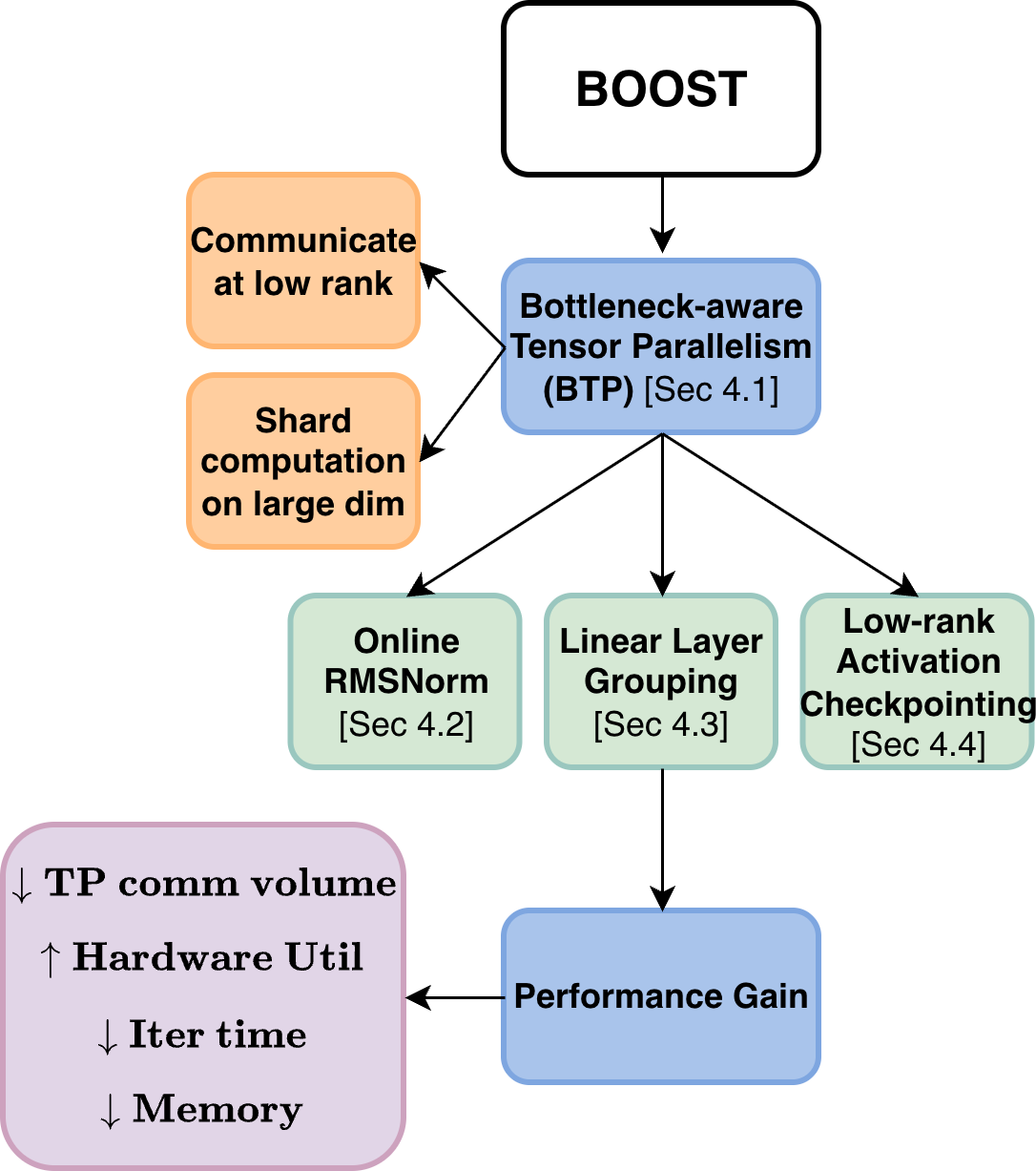}\\
  \end{minipage}
  \caption{(Left) Linear layer in Bottleneck Architecture; (Middle) Decoder block runtime breakdown among different TP strategy; (Right) Overview of our framework \framework.}
  \label{fig:intro}
  \vspace{-10pt}
\end{figure*}

\section{Introduction}
\label{sec:introduction}
Large Language Models (LLMs) continue to evolve and are increasingly becoming an integral part of a wide range of industrial and scientific applications. More generally, modern transformer-based foundation models that combine multi-modal data, use domain-specific
information, and being integrated into complex AI workflows are pushing the limits of scientific 
discovery. Such models become more useful as they grow larger and ingest more training data ~\cite{kaplan2020scaling, hoffmann2022training}. Unfortunately, this unprecedented scale of
growth also poses a major challenge: pre-training such models is exceptionally costly.
For example, GPT-3 (175B) was trained on roughly 300B tokens~\cite{brown2020language}; BLOOM-176B required 1.08M A100 GPU-hours~\cite{luccioni2023estimating}; Meta’s LLaMA-3.1-405B model was trained on 15T tokens with 30.84M H100 GPU hours~\cite{dubey2024llama}. As the trend continues, it takes hundreds of millions of US dollars to develop such a state-of-the-art foundation model. For example, the cost of developing Grok-4 is estimated to be $\sim$\$500M~\cite{epoch2025grok4trainingresources}. These realities motivate both academia and industry to actively develop efficient training techniques for foundation models.

\paragraph*{\bf Motivation:} Driven by the need of scalability, major state-of-the-art advancements have targeted both 
system-level and algorithmic considerations. From the system-level perspective, a key challenge
is how to distribute the computations and data on multiple GPUs in order to parallelize computations and aggregate the GPU memory (and other memory tiers) while reducing coordination costs and communication overheads. In this regard, techniques such as 3D parallelism (a combination of data, tensor/sequence, and pipeline)~\cite{shoeybi2019megatron, korthikanti2023reducing, huang2019gpipe, narayanan2019pipedream, qi2023zero} are indispensable in modern training frameworks such as DeepSpeed~\cite{rasley2020deepspeed}, Nanotron~\cite{tazi2025ultra} and TorchTitan~\cite{liang2024torchtitan}. Reduced communication
costs and coordination overheads are often achieved by using high-performance communication libraries (e.g., NCCL and MPI~\cite{walker1996mpi, gabriel2004open}) and asynchronous I/O techniques that hide the cost of processing messages and GPU-host memory movements in the background. From the algorithmic perspective, many efforts have sought to improve model efficiency through different directions, including improved attention mechanisms~\cite{wang2020linformer, shazeer2019fast, ainslie2023gqa, yuan2025native}, quantization-based methods that preserve accuracy under reduced precision~\cite{micikevicius2022fp8, peng2023fp8, chmiel2025fp4, abecassis2025pretraining, xi2023training}, and sparsity-based approaches such as Mixture-of-Experts (MoE)~\cite{jacobs1991adaptive, jordan1994hierarchical}.

Among these algorithmic approaches, low-rank methods that either project gradients/activations to low-dimensional space~\cite{zhao2024galore, zhu2024apollo, shamshoum2024compact, chen2025memory} or have partial/entire weights factorized in low-rank matrix/tensor format~\cite{lialin2023relora, loeschcke2024loqt, han2024sltrain, yang2024comera, liu2025cola, LORO_iclr2025, zhang2025lax, kong2025cr, li2025lost} are a particular class of algorithmic methods specifically targeted at balancing accuracy with performance and resource utilization. Thanks to a unified bottleneck architecture (Figure~\ref{fig:intro} left), examples such as CoLA~\cite{liu2025cola}, LORO~\cite{LORO_iclr2025} and LaX~\cite{zhang2025lax} can simultaneously reduce parameter count, memory, compute and runtime compared with a full-rank architecture while preserving comparable accuracy. Unfortunately, a large part of these efforts have only been explored at small scale (e.g., below 7B parameters), in which case both the full-rank and the corresponding low-rank model can fit in the memory of a single GPU. Without a co-design with system-level approaches, 
the scalability potential of low-rank methods remains limited. Specifically, one difficult challenge is how to enable efficient tensor parallelism for low-rank methods, which, unlike data and pipeline parallelism, features a tightly coupled pattern with all-to-all dependencies. In this paper, we aim to address this challenge.

{\bf Limitations of state of the art:}
We cannot simply apply vanilla tensor parallelism techniques that were originally designed for full-rank methods to low-rank methods. As can be
observed in Figure~\ref{fig:intro} (middle), a full-rank training iteration 
on four GPUs co-located on the same compute node (details in \S~\ref{sec:experimental-setup}) has less than 20\% communication overhead,
while an equivalent low-rank method experiences an explosion in the communication
overhead. This is because the bottleneck architecture is based on smaller but deeper
structures that feature more synchronization points, which increase
the communication latency and volume. Additionally, the GPU utilization is also sub-optimal due to inefficient shapes of the weight matrices as partitioned by vanilla tensor parallelism onto individual devices, leading to insufficient low-rank computational speed-up that erodes algorithmic efficiency. Together, these limitations restrict the theoretical algorithmic scalability potential.

{\bf Contributions:} To bridge this gap, we design and implement scalable tensor-parallel training strategy for efficient pre-training of low-rank (bottleneck) foundation models. We leverage the idea that synchronization points can either be avoided or placed strategically where the bottleneck architecture is narrow to reduce communication overheads. And we boost computation by sharding along the large dimensions to preserve healthy GEMM reduction sizes thus improving GPU utilization.
We summarize our key contributions as follows:
\begin{itemize}[leftmargin=*, noitemsep, topsep=0pt]
\vspace{-5pt}
  \item We present a theoretical analysis of bottleneck architectures in 3D parallelism, quantifying arithmetic intensity and communication volume in distributed training to reveal the scaling challenges of vanilla designs and the benefits of our approach.
  \item We propose a novel tensor parallelism strategy, namely, {\it Bottleneck-aware Tensor Parallelism} (BTP), which optimizes partitioning of low-rank weight matrices to facilitate efficient communication on low-rank activations. BTP also significantly increases the computational intensity of GEMM kernels, thus improving GPU utilization.   
  \item We implement \framework, a high-performance distributed training framework to take advantage of BTP in several ways: {\it Online-RMSNorm} to facilitate sharded-safe global normalization and reduce latency, {\it Layer Grouping} to improve computational intensity and lower the number of collective operations, and {\it Low-rank Activation Checkpointing} to reduce re-computation cost and eliminate additional collectives. 
  \item We demonstrate a 1.46–1.91$\times$ speedup over full-rank baselines and a 1.87–2.27$\times$ speedup over vanilla tensor parallel implementations of bottleneck architectures. The ablation study further confirms efficiency gains on both the computation and communication axes.

\end{itemize}

\section{Related Work}
\label{sec:related}

Low-rank methods have been extensively explored in the literature from different perspectives, the most relevant of which we discuss below.

{\bf Low-Rank Matrix Factorization.} Classical low-rank training schemes~\cite{khodak2021initialization, kamalakara2022exploring} replace a weight matrix $\mat{W} \in \mathbb{R}^{d_{\text{out}} \times d_{\text{in}}}$ with $\widehat{\mat{W}}\approx \mat{BA}$, where $\mat{A} \in \mathbb{R}^{r \times d_{\text{in}}}$, $\mat{B} \in \mathbb{R}^{d_{\text{out}} \times r}$, $r \ll \min(d_{\text{out}}, d_{\text{in}})$ determines the rank and thus the compression ratio. However, naively applying this low-rank matrix factorization often harms model accuracy in pre-training setting. Recent work proposed various modifications to improve model capacity. For example,  
  ReLoRA~\cite{lialin2023relora} adopted the LoRA~\cite{hu2022lora} adapter with constantly merged and reinitialized low-rank factors.
 SLTrain~\cite{han2024sltrain} combined an unstructured sparse matrix with low-rank factors to approximate $\mat{W}$.
 LORO~\cite{LORO_iclr2025} preserved the classical low-rank formulation but perform some optimization steps directly on the Riemannian manifold. 

{\bf Low-Rank Activation via Auto-Encoder.} Observing that the activation function of every layer stays in a low-rank space, CoLA \cite{liu2025cola} proposed to replace both an MLP layer  $\sigma(\mat{Wx)}$ and a linear projection layer $\mat{Wx}$ by an auto-encoder layer $\mat{B}\sigma(\mat{Ax})$. This method has achieved simultaneous reduction of perplexity, model size, memory cost and runtime in pre-training BERT and LLaMA. Recently LOST~\cite{li2025lost} further modified CoLA by incorporating an unstructured sparse matrix $\mat{S}$, similar to the ones in SLTrain, as $\alpha \mat{B}\sigma(\mat{Ax}) + (1-\alpha) \mat{Sx}$.

{\bf Low-Rank Models Enhanced by Residual Connection.} Another promising direction is to boost the accuracy of existing low-rank architecture with some plug-in residual blocks. LaX~\cite{zhang2025lax} proposed to fuse cross-layer low-rank features to boost model capacity without increasing their physical rank, i.e., $\mat{B}_i(\mat{A}_i\mat{x}_i + \mat{A}_{i-1}\mat{x}_{i-1})$. This method can improve the accuracy of low-rank matrix/tensor-compressed models as well as CoLA with negligible parameter and computing overhead. CR-Net~\cite{kong2025cr} proposed a similar approach from a different perspective: the difference of activations between adjacent layers exhibit low-rank property, thus $\mat{B}_i\mat{A}_i\mat{x}_i + \mat{B}_{i-1}\mat{A}_{i-1}\mat{x}_{i - 1}$, with the exception of the first layer being full-rank.


Additionally, pre-training LLMs also requires sophisticated large-scale system-level optimizations. Two complementary paradigms dominate large-scale training: 

{\bf Redundancy Elimimation using Model/Optimizer State Sharding.}
Methods such as ZeRO~\cite{rajbhandari2020zero}, ZeRO-Offload~\cite{ren2021zero}, ZeRO-Infinity~\cite{rajbhandari2021zero} and PyTorch FSDP~\cite{zhao2023pytorch} partition optimizer states, gradients, and parameters across devices to reduce memory footprint and enable training larger models. Representative implementations include DeepSpeed ZeRO~\cite{rasley2020deepspeed} and FairScale’s sharded optimizers~\cite{FairScale2021}. Such techniques are complementary to our
own work.

\begin{figure}[t]
 \centering
 \includegraphics[width=0.5\textwidth]{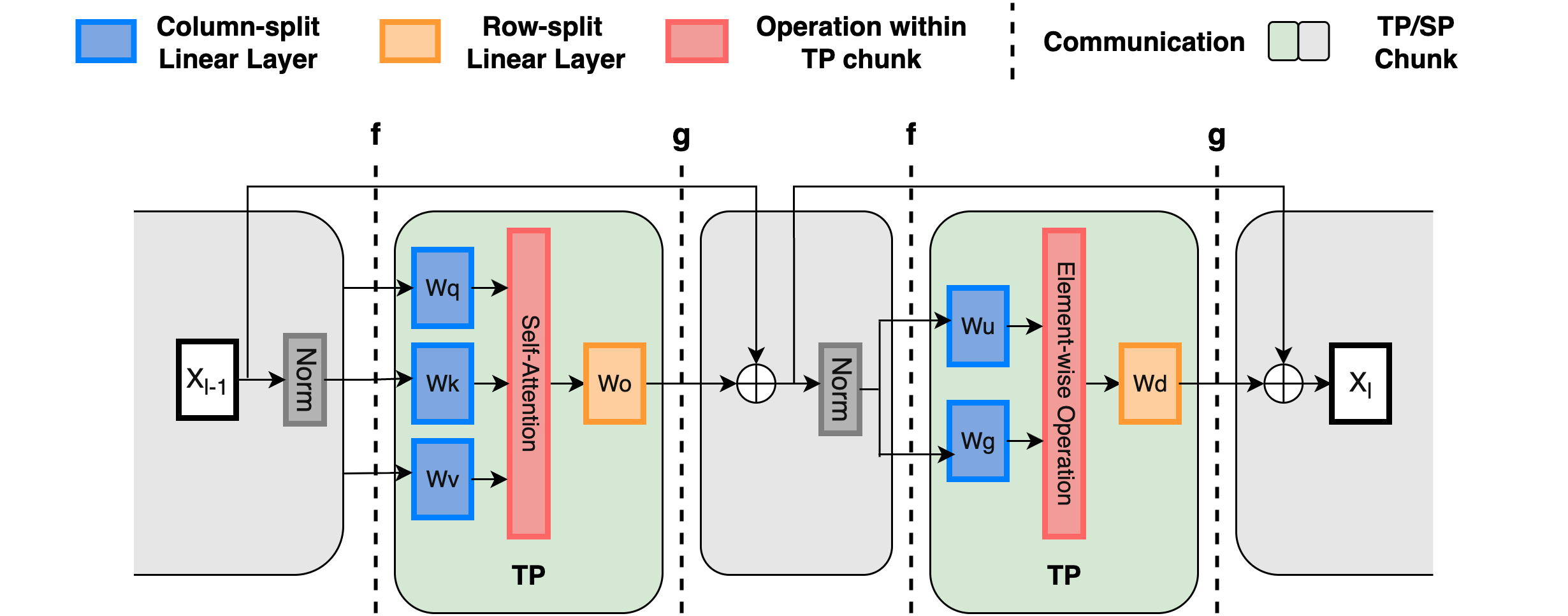}
 \caption{Megatron-LM style Tensor Parallelism. Here, $f$ is identity in forward and all-reduce in backward, while $g$ is all-reduce in forward and identity in backward.}
 \label{fig:megatron TP}
 \vspace{-10pt}
\end{figure}

\textbf{\bf 3D Parallelism.}
Frameworks such as Megatron-LM and Megatron-DeepSpeed combine data (DP), tensor (TP), and pipeline (PP) parallelism to distribute memory and computation across devices for billion-parameter pre-training \cite{shoeybi2019megatron,narayanan2021efficient}. DP accelerates training by replicating the model across GPUs, processing different mini-batches in parallel, and synchronizing by \emph{gradient all-reduce} once per iteration before the optimizer step. PP splits layers vertically into stages so micro-batches stream through as an assembly line, communicates activations/gradients at stage boundaries. Efficiency hinges on schedules such as GPipe/PipeDream \cite{huang2019gpipe,narayanan2019pipedream} and on minimizing warm-up/drain bubbles. TP partitions the model horizontally, sharding individual weight matrices and corresponding activations across GPUs. Each device computes partial results that are later combined via tightly-coupled collective communication. TP builds on matrix/activation sharding (e.g., Mesh-TensorFlow, GShard, GSPMD) \cite{shazeer2018mesh,lepikhin2020gshard,xu2021gspmd}. Megatron-LM adopts a standard \emph{column--row} pattern: each TP \emph{chunk} (two linears plus the intervening op such as SDPA/SwiGLU) issues one collective per pass. Here, $f$ is identity in the forward pass and all-reduce in the backward pass, while $g$ is all-reduce in the forward pass and identity in the backward pass (Figure~\ref{fig:megatron TP}) \cite{shoeybi2019megatron}. 

\textbf{\bf Tensor Parallelism in Hybrid MoE Systems.} Mixture-of-Experts (MoE) has become a major trend in modern LLM scaling. Although expert parallelism (EP) is a common scaling strategy in MoE systems, it introduces heavy token-level all-to-all communication and can suffer from severe load imbalance. By contrast, tensor parallelism (TP) relies on more regular collective communication, often within a node. Moreover, when experts become sufficiently large, EP may complement rather than replace TP, and hybrid EP+TP configurations can still be necessary~\cite{singh2023hybrid, jin2025megascale, liu2025moe}. As a result, TP remains important in modern large-model training, including hybrid dense--MoE systems: large dense/shared components still rely on TP, and sufficiently large experts may also require TP in addition to EP.


\textbf{Tensor Parallelism for Low-Rank Adapters.} Prior work on low-rank methods has focused more on fine-tuning than on pre-training. S-LoRA~\cite{sheng2024slora} proposes a tensor-parallel strategy for the low-rank adapter setting, where LoRA adapters are attached to a frozen dense backbone. While the adapter branch also introduces deeper layers with narrow intermediate activations, its output must be merged back into the frozen full-rank branch, so the communication and sharding design remains naturally coupled to the dense backbone. In contrast, our work focuses on pre-training pure low-rank or bottleneck backbone architectures without a frozen full-rank branch.


To the best of our knowledge, we are the first to tackle the problem of optimizing tensor parallelism for low-rank methods in pre-training. In this work, we focus on \textbf{dense} low-rank bottleneck architectures, which are the main setting considered by current low-rank \textbf{pre-training} methods, and design a general framework for unified bottleneck architectures such as CoLA, LORO, and LaX. In practice, this class of methods typically offers the largest training speedups and memory savings with minimal impact on accuracy. Furthermore, although outside the scope of this work, our framework is complementary to pipeline and data parallelism, as well as other techniques such as redundancy elimination; we discuss broader applicability to architectures such as MoE in Sec.~\ref{sec:discussion}.

\section{Preliminaries}
\label{sec:challenges-principles}

As shown in Figure~\ref{fig:intro} (Middle), naively scaling a bottleneck architecture exhibits both \emph{communication} and \emph{computation} inefficiencies. We quantify system efficiency with two complementary metrics:
\begin{itemize}[leftmargin=*,noitemsep, topsep=0pt]
    \item {\bf Communication Volume} $V_{\text{comm}}$: bytes moved by collectives, lower $V_{\text{comm}}$ indicates reduced synchronization overhead and better scalability across devices;     
    \item {\bf Arithmetic Intensity (A.I.)}: FLOPs per byte moved by kernels, a proxy for how close execution is to compute-bound. A higher A.I. indicates a compute-bound operation capable of saturating GPU compute units, while a lower A.I. denotes a memory-bound kernel limited by memory bandwidth. 
For a matrix multiplication $\mat{C} = \mat{A} \times \mat{B}$ with $\mat{A} \in \mathbb{R}^{M \times K}$, $\mat{B} \in \mathbb{R}^{K \times N}$, and $\mat{C} \in \mathbb{R}^{M \times N}$, the total number of FLOPs is $2MNK$. Assuming ideal memory reuse, the total memory traffic consists of reading $\mat{A}$ and $\mat{B}$ and writing $\mat{C}$, yielding:
\begin{equation}
\text{A.I.} = \frac{2MNK}{(MK + KN + MN) \cdot \text{bytes\_per\_element}}.
\label{eq:ai-matmul}
\end{equation}
\end{itemize}

\textbf{DP and PP in Bottleneck Architectures.} Table~\ref{tab:perf-analysis-table} summarizes the communication volume across different parallelism strategy. Low-rank bottleneck architectures \emph{intrinsically} benefit DP and PP due to smaller parameter size and low-rank activations. Because factorized weights yield smaller gradients, bottleneck models reduce gradient all-reduce traffic. In LLaMA with $r=\tfrac{d}{4}$, this corresponds to an $\sim\!2.5\times$ reduction in communication volume. For pipeline parallelism, factorized layers lower per-stage FLOPs, which shortens per micro-batch compute and shrinks the pipeline bubble; the smaller footprint also permits larger micro-batches, improving arithmetic intensity and hardware utilization.

\begin{table}[t]
\caption{
Per-iteration communication volume ($V_{\text{comm}}$) under different parallelism strategies for full-rank and bottleneck architecture in LLaMA-like models.
}
\label{tab:perf-analysis-table}
\vskip 0.05in
\begin{center}
\begin{small}
\begin{sc}
\resizebox{\linewidth}{!}{%
\begin{tabular}{lccc}
\toprule
Strategy & Full-Rank & \multicolumn{2}{c}{Low-Rank} \\
         &           & Vanilla & BOOST \\
\midrule
Data Parallelism & $O(d(d+d_{ff}))$ & \multicolumn{2}{c}{$O(r(d+d_{ff}))$} \\
Pipeline Parallelism & $O(d)$ & \multicolumn{2}{c}{$O(d)$} \\
Tensor Parallelism & $O(d)$ & $O(d+d_{ff})$ & $O(r)$ \\
\bottomrule
\end{tabular}
}
\end{sc}
\end{small}
\end{center}
\vskip -0.1in
\end{table}

{\bf TP in Bottleneck Architectures.} Contrary to being mostly model-agnostic in DP and PP, TP interacts directly with model structure: it shards activations and parameters in architecture-aware ways to balance compute intensity and communication. In the following section we will analyze why a naive low-rank TP degrades efficiency and motivates our bottleneck-aware TP design.

\section{\framework: BOttleneck-Optimized Scalable Training}
\label{sec:design}

In this section, we introduce our \framework\ framework and present the details of how it addresses scalability issues of low-rank (bottleneck) LLM training under 3D parallelism. 

\begin{figure*}[t]
  \centering
  \includegraphics[width=\textwidth]{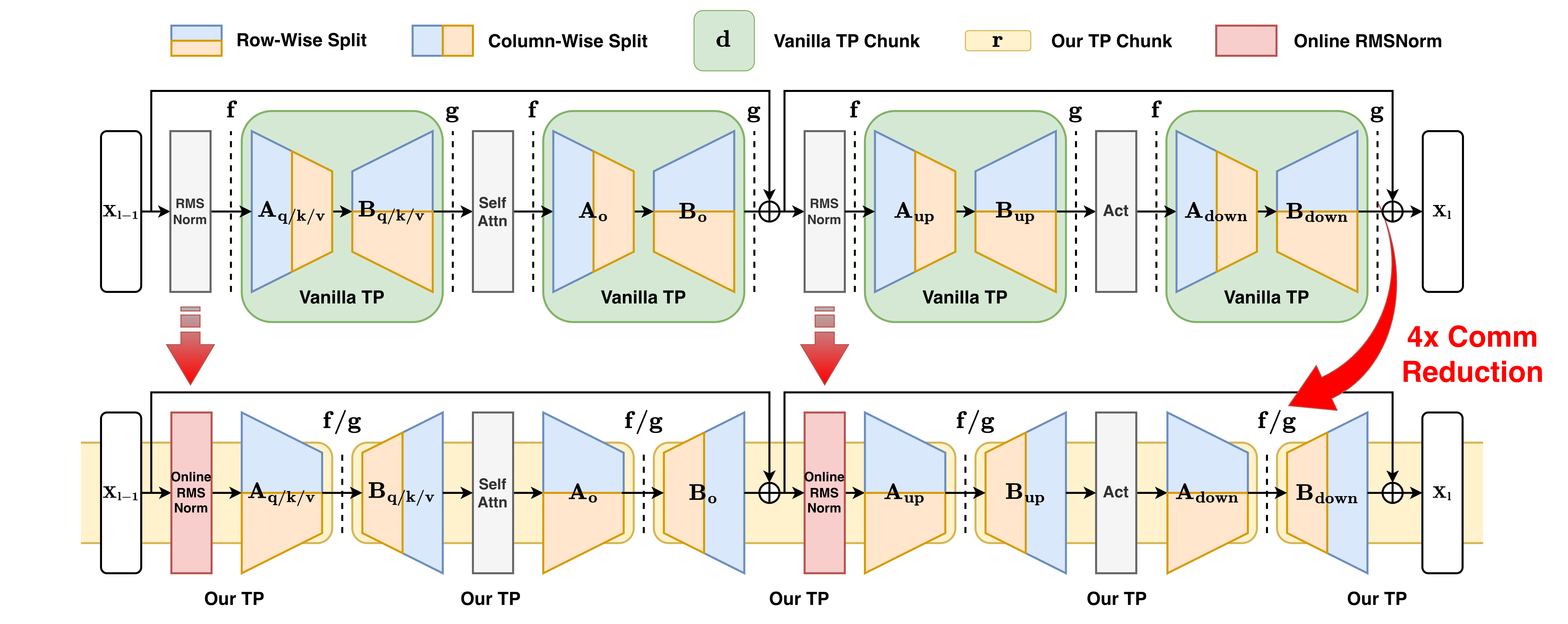}
  \vspace{-20pt}
  \caption{Modularized Tensor Parallelism Design for Bottleneck Architecture. Top: vanilla TP uses separate $f$ and $g$ communication boundaries for each TP chunk. Bottom: after shifting the TP chunk boundary to the low-rank bottleneck, the same low-dimensional boundary serves as a shared $f/g$ point.}
  \label{fig:BTP}
  \vspace{-10pt}
\end{figure*}

\subsection{Bottleneck-aware Tensor Parallelism}
\label{sec:bottleneck-tp}

Tensor parallelism is non-trivial for bottleneck layers. Replacing a single $d{\times}d$ projection with consecutive $d{\times}r$ and $r{\times}d$ projections ($r\!\ll\! d$) \emph{doubles} linear stages per block and raises two design questions: (i) \emph{Where to place TP chunks in deeper block?} Since each TP chunk incurs one collective, adding stages increases the number of collectives per block. (ii) \emph{How to shard efficiently given the low-rank structure?} Conventional TP reduces per-rank hidden width, but the bottleneck factorization already shrinks effective dimension, inefficient sharding can therefore harm hardware utilization by having a lower arithmetic intensity.

\textbf{Vanilla low-rank Tensor Parallelism.} A straightforward baseline follows the Megatron-LM pattern: treat each pair of low-rank layers
($d{\times}r$ then $r{\times}d$) as a TP chunk: column-parallel for the down-projection, row-parallel for the up-projection. We refer to this as \emph{vanilla TP} (top of Figure~\ref{fig:BTP}). However, this vanilla design introduces both communication and computation inefficiencies. 

On the \emph{communication} side, a full-rank Megatron style decoder block issues two activation-sized collectives of payload $[b,s,d]$, i.e., $V_{\text{comm}}^{\text{full}}=2bsd$ (one in attention, one in MLP). In contrast, vanilla TP triggers a collective at \emph{each} linear: $4\,bsd$ in the attention block and $bsd+2\,bs\,d_{ff}$ in the MLP, for a total of 
\begin{equation}
V_{\text{comm}}^{\text{vanilla}} \;=\; (5\,bsd + 2\,bs\,d_{ff})
\;=\; \Big(\tfrac{5 + 2(d_{ff}/d)}{2}\Big)\times V_{\text{comm}}^{\text{full}}. 
\label{eq:btp_communication_volume}
\end{equation}
Thus the per-block volume grows by $\mathbf{5\times}$ when $d_{ff}{\approx}2.5d$, and up to $\mathbf{6.5\times}$ when $d_{ff}{\approx}4d$. This explains the large communication time observed in the breakdowns for naively scaled bottleneck models.

On the computation side, bottleneck architecture already reduces the effective hidden dimension, and vanilla TP further shards along the low rank $r$, shrinking the GEMM reduction dimension. This results in kernels that move a lot of data but perform little compute, lowering arithmetic intensity and pushing execution into the memory-bound regime. Although vanilla low-rank TP reduces FLOPs, its A.I. remains very low due to disproportionate data movement; for example, in LLaMA-7B MLP blocks it attains only 
$0.2\times$ the A.I. of full-rank TP, leading to poor GPU utilization. This also explains why at scale the reduction in computation time is smaller than in the single-GPU setting.

Additionally, the up-projection materializes a full, unsharded activation even though each TP rank consumes only its local shard, leading to redundant allocation and unnecessary data movement.

\textbf{Bottleneck-aware Tensor Parallelism (BTP).} To address the inefficiencies of vanilla TP, we propose a bottleneck-aware sharding strategy that leverages the low-rank $r$-dimensional output of each pair of bottleneck layers to issue lightweight collectives. Concretely, this sharding strategy shifts the TP chunk boundary by exactly one individual bottleneck layer. The bottleneck-aware TP chunk starts with the \emph{up-projection} layer ($r\times d$) being \emph{column}-parallel and the \emph{next} \emph{down-projection} layer ($d\times r$) being \emph{row}-parallel, while the intervened operations being executed on sharded activations. To shard along the large dimension and communicate at the low dimension, {\bf BTP improves both the communication efficiency and computation efficiency.}

Effectively, BTP improves communication efficiency by reducing its workload. In BTP chunk, the collective operates on a \emph{low-rank} activation of size $[b,s,r]$ rather than $[b,s,d]$. Per decoder block, this yields a total payload of 
\begin{equation}
V_{\text{comm}}^{\text{BTP}} \;=\; 7\,bsr
\;=\; \Big(\tfrac{7r}{2d}\Big)\times V_{\text{comm}}^{\text{full}}.
\label{eq:btp_communication_volume}
\end{equation}
Thus BTP reduces communication volume by more than $5.7\times$ versus vanilla low-rank TP (when $r=\frac{d}{4}$) and by $1.14\times$ versus the full-rank TP baseline, substantially lowering communication time in practice.

From the computation side, BTP shards along the hidden dimension rather than the low-rank dimension $r$, producing kernels with a larger GEMM reduction dimension and thus higher arithmetic intensity. Although BTP and vanilla TP perform the same FLOPs, BTP moves less data, yielding higher arithmetic intensity. For example, in LLaMA-7B MLP blocks, BTP achieves $2.5\times$ the A.I. of vanilla TP, translating into substantially better hardware utilization.




\subsection{Online RMSNorm}
\label{sec:online-rmsnorm}

\textbf{Sharded-safe vs.\ sharded-unsafe operators.}
In tensor parallelism, we assume no extra collectives inside a TP chunk: a single all-reduce is issued at the end of each chunk.
Under this assumption, any operation inside the chunk that can be correctly performed on sharded activations is \textbf{sharded-safe}, e.g., element-wise functions (activations, dropout) and attention heads when the head count is divisible by the TP degree.
Conversely, ops that require cross-shard data are \textbf{sharded-unsafe}; a canonical example is normalization, which depends on global mean and variance computations and therefore cannot be placed inside a TP chunk without additional synchronization.

A simple design principle is to avoid sharded-unsafe operations inside the chunk. However, with BTP the RMSNorm [Eq.~\eqref{eq:rmsnorm}] naturally falls in the middle of a TP chunk, so we must handle it explicitly. We therefore introduce and compare two variants: Sync RMSNorm (explicit in-place statistic synchronization) and Online RMSNorm (fused statistic exchange with recovery).
\begin{equation}
\begin{aligned}
\text{RMSNorm}(\mat{X}) &= \frac{\mat{X} \cdot \boldsymbol{\gamma}}{\text{RMS}(\mat{X})}, \\
\text{with} \quad \text{RMS}(\mat{X}) &= \sqrt{\frac{1}{d} \sum_{j=1}^{d} \mat{X}_{:,j}^2 + \epsilon}.
\normalsize
\end{aligned}
\label{eq:rmsnorm}
\end{equation}
\textbf{Sync RMSNorm.} A straightforward approach is to compute per-rank local statistics $\text{RMS}(\mat{X}_i)$ and use a collective to aggregate the global statistic before normalizing. Although the payload is tiny (the statistic exchange at $[b,s,1]$), many such calls incur nontrivial launch overhead, are latency-dominated and achieve poor effective bandwidth, resulting in extra communication time.

\textbf{Online RMSNorm.}
To eliminate the standalone small-payload collective, we compute normalization using local statistics and fuse the statistic exchange into the TP collective that follows the next GEMM. This mirrors the insight of online-softmax in FlashAttention~\cite{dao2022flashattention}: defer global normalization and recover the exact result afterward. In effect, the previously sharded-unsafe RMSNorm becomes a sharded-safe online RMSNorm.


\begin{algorithm}[tb]
   \caption{Online RMSNorm}
   \label{alg:online-rmsnorm}
   \textbf{Input:} $d_{\text{local}} = d / \text{TP\_size}$; sharded activation $X_i \in \mathbb{R}^{s \times b \times d_{\text{local}}}$ on TP rank $i$; row-split weight $W_i \in \mathbb{R}^{d_{\text{local}} \times r}$; scaling factor $\gamma_i$; numerical constant $\epsilon$ \\
   \textbf{Output:} Final output $Y \in \mathbb{R}^{s \times b \times r}$
   \begin{algorithmic}[1]
      \STATE $S_\text{local} \gets \sum_{j=1}^{d_{\text{local}}} X_i[:, :, j]^2$
      \STATE $rms_{\text{local}} \gets \sqrt{\frac{1}{d_{\text{local}}} S_{\text{local}} + \epsilon}$
      \STATE $X_i \gets \frac{X_i}{rms_{\text{local}}} \times \gamma_i$
      \vspace{2mm}
      \STATE $H_i \gets X_i \cdot W_i$
      \STATE $H_i \gets H_i \cdot rms_{\text{local}}$
      \STATE $[H_{\text{global}}, S_{global}] \gets \text{AllReduce}([H_i, S_{local}])$
      \STATE $rms_{\text{global}} \gets \sqrt{\frac{1}{d} S_{\text{global}} + \epsilon}$
      \STATE $Y \gets H_{\text{global}} \cdot \frac{1}{rms_{\text{global}}}$
      \STATE \textbf{return} $Y$
   \end{algorithmic}
\end{algorithm}

The online-RMSNorm operator consists of three main steps (Alg.~\ref{alg:online-rmsnorm}):
\textbf{Step 1: Local norm computation.} Compute and record the local sum of squares on sharded activations (Line 1) and calculate local RMSNorm (Line 2 and 3).
\textbf{Step 2: Row-split GEMM with fused all-reduce.} Apply the following row-parallel GEMM (Line 4), rescale the rank-dependent correction factor (Line 5), then all-reduce together the GEMM output and the local statistic (Line 6).
\textbf{Step 3: Recovery.} Calculate correct global rms using global sum of squares (Line 7). And rescale the reduced GEMM output back by the global rms(Line 8).

Online RMSNorm preserves \textbf{mathematical equivalence}: although the GEMM is performed using locally normalized inputs, the output can be rescaled by a per-row correction factor derived from the ratio of local to global statistics, making the result equivalent to standard RMSNorm. Specifically, we can show in Equation~\eqref{eq:online-rmsnorm-correctness}:
\begin{equation}
\sum_i(\frac{\mat{W_i}\mat{X_i}}{\text{RMS}(\mat{X})}) = (\sum_i  \frac{\mat{W_i} \mat{X_i}}{\text{RMS}(\mat{X_i})} \cdot \text{RMS}(\mat{X_i})) \cdot \frac{1}{\text{RMS}(\mat{X})}
\label{eq:online-rmsnorm-correctness}
\end{equation}

\begin{figure}[t]
  \centering
  \includegraphics[width=0.5\textwidth]{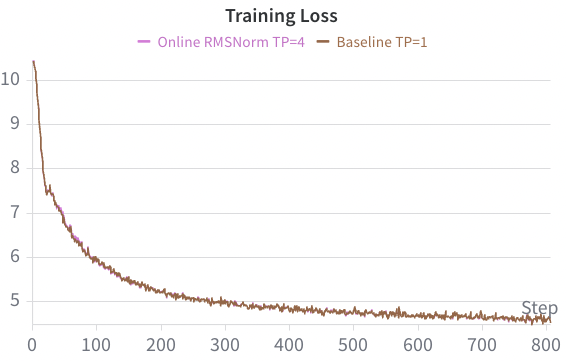}
  \vspace{-20pt}
  \caption{Training loss curves for a tiny LLaMA model, comparing BTP with Online RMSNorm against the TP=1 baseline. The two curves closely match, indicating that the corrected Online RMSNorm preserves stable training behavior.}
  \label{fig:Online_RMSNorm_Training_Curve}
  \vspace{-10pt}
\end{figure}

Moreover, Online RMSNorm helps maintain \textbf{numerical stability} in practice. By computing intermediate normalization statistics locally on sharded activations, it reduces the likelihood of extreme values that could otherwise lead to overflow or underflow. Online RMSNorm can closely match the $TP=1$ baseline at both the kernel and training levels. As shown in Table~\ref{tab:online_rmsnorm_equiv}, for $TP=4$, Online RMSNorm followed by row-split linear yields only small numerical discrepancies from the $TP=1$ baseline in both fp32 and bf16, within acceptable precision tolerances. Fig.~\ref{fig:Online_RMSNorm_Training_Curve} further shows that BTP with Online RMSNorm preserves the baseline training behavior.

\begin{table}[t]
\centering
\small
\caption{Kernel-level comparison between Online RMSNorm + row-split linear (TP=4) and the TP=1 baseline RMSNorm + linear output. We report the average maximum and mean absolute differences.}
\vspace{0.5em}
\label{tab:online_rmsnorm_equiv}
\begin{tabular}{lcc}
\toprule
\textbf{Precision} & \textbf{Avg. Max Abs. Diff.} & \textbf{Avg. Mean Abs. Diff.} \\
\midrule
FP32 & $7\times10^{-7}$ & $6\times10^{-8}$ \\
BF16 & $3.125\times10^{-2}$ & $2.2\times10^{-3}$ \\
\bottomrule
\end{tabular}
\end{table}

From a \textbf{performance standpoint}, local normalization reduces per-rank work via smaller dimension, and fusing the statistic exchange with the GEMM all-reduce removes a separate kernel launch and increases effective bandwidth by aggregating it into a larger message.
The recovery is a cheap per-row scaling.
Empirically, the recovery compute overhead is small relative to the latency-domiated statistic collective in Sync RMSNorm, making the computation-communication trade-off favorable.

In practice, these two variants offer different trade-offs. \textbf{Sync RMSNorm} is the exact and conservative option: it explicitly synchronizes the scalar normalization statistic before applying RMSNorm, and therefore serves as a simple always-available fallback. \textbf{Online RMSNorm} uses a reordered computation to defer synchronization and recover the same global normalization factor after the following TP collective, making it the faster option when minimizing standalone small-payload communication is important.

\subsection{Linear Layer Grouping}
\label{sec:grouping}
Although BTP outperforms vanilla TP, its extra sync point and small GEMMs still degrade efficiency. To translate BTP’s analytical gains into consistent per-layer speedups, we apply linear-layer grouping optimization.

In a full-rank model, we group parallel linears into a single fused operation (e.g., QKV in attention; gate+up in MLP) to reduce kernel launches and enlarge GEMMs. However, grouping is more challenging for bottleneck layers because branch inputs differ: while down-projections share $\mat{X}$ and can be grouped by concatenating weights, up-projections use distinct inputs ($\mat{X}_1,\mat{X}_2,\mat{X}_3$), so we use a batched GEMM over (input, weight) pairs in one fused kernel.

Under BTP, grouping delivers even larger benefits. On the compute path, it cuts kernel count and eliminates redundant reads of $X$, reducing total data movement, raising arithmetic intensity (A.I.) and hardware utilization. On the communication path, it lowers the number of collectives and improves effective bandwidth by increasing per-call payload, together producing a substantial boost in end-to-end efficiency.

\begin{figure}[t]
 \centering
 \includegraphics[width=0.98\linewidth]{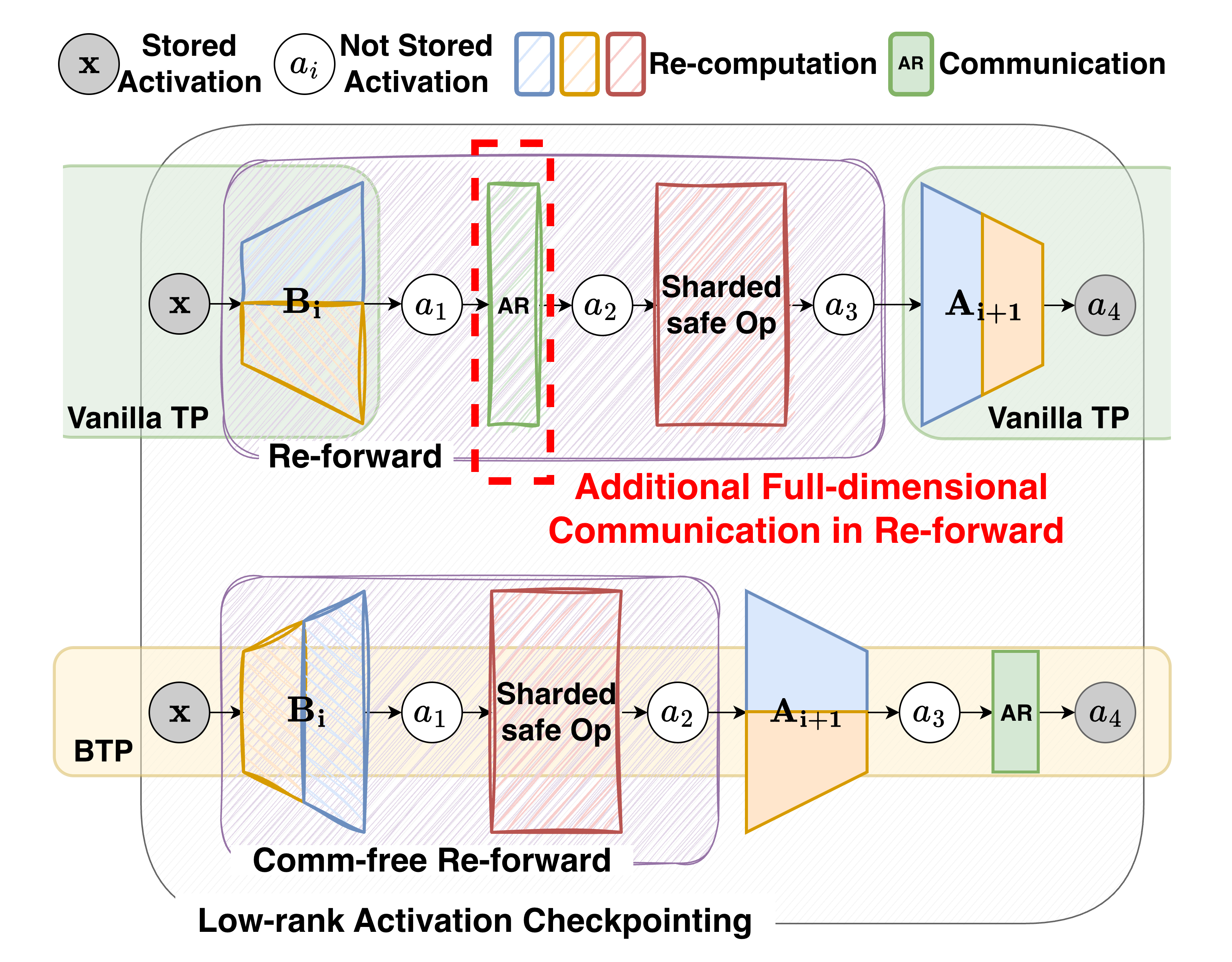}
 \caption{Low-rank efficient activation checkpointing under BTP}
 \label{fig:lr_activation_checkpointing}
 \vspace{-10pt}
\end{figure}

\subsection{Comm-free Low-rank Activation Checkpointing}
\label{sec:activation checkpointing}
As the model scales up, we use PP to span across nodes. However, standard 1F1B scheduling requires multiple microbatches in flight to fill pipeline bubbles, which inflates activation memory. Activation checkpointing is a common remedy, where intermediate results are partially saved in forward, and re-computed when needed in backward. When integrating with the bottleneck architecture, its low-rank activations provide a better choice for choosing the checkpointing interval that favors the tradeoff between memory and compute, as exemplified by CoLA-M~\cite{liu2025cola}.

However, vanilla-TP would weaken such benefit by introducing additional communications. From Figure~\ref{fig:lr_activation_checkpointing}, we see that its re-forward path contains the collective operation that lives between two consecutive TP chunks. Effectively, this triggers an extra synchronization point during re-forward.

In contrast, our BTP amplifies the benefit of low-rank checkpointing by aligning its boundaries with the checkpoint interval: the re-forward stays \emph{entirely within-chunk} and requires no additional communication. As shown in Figure~\ref{fig:lr_activation_checkpointing}, within each BTP chunk (yellow) we checkpoint only low-rank activations ($x$ and $a_4$). In backward, we locally re-forward $x \rightarrow B_i \rightarrow a_1 \rightarrow \text{sharded-safe} \rightarrow a_2$ to reconstruct the input needed for the gradient computation of the subsequent up-projection $\mat{A}_{i+1}$, resulting in a completely \emph{communication-free} re-forward.

\subsection{Implementation}
\label{sec:impl}
\textbf{Software Stack.} 
We integrate \framework into Nanotron~\cite{tazi2025ultra} pre-training framework as composable engines. Nanotron provides column/row-parallel linear APIs, a 3D (DP/TP/PP) execution engine, NCCL-based collectives that run on a dedicated communication stream, with built-in activation checkpointing. In our setup we operate in all-reduce mode (no sequence-parallel chunks), and Nanotron currently do not support fine-grained overlapped all-reduces.

\textbf{Model Implementation.}
For full-rank baseline we use Nanotron’s reference model. Low-rank variants factorize $d{\times}d \!\to\! d{\times}r,\, r{\times}d$ and select matching col/row-parallel linears. For BTP, we shard the embedding output to row-split the first down-projection layer; the final up-projection layer is replicated on every TP rank. Online RMSNorm computes on sharded activations, returns local RMS, and piggybacks local stats in the next TP collective via \textit{all\_reduce\_coalesced}. Grouping adds (i) concatenated down-projections (shared $X$) and (ii) batched-GEMM up-projections (distinct inputs) to cut launches and traffic. Checkpointing stores only low-rank boundaries via Nanotron’s decorator.

\textbf{Configuration.}
Building on Nanotron’s example configs, we expose minimal knobs to toggle \framework components. When preconditions fail, we fall back to unfused operators or Sync RMSNorm; unsupported cases are auto-disabled safely.

\textbf{Profiling.}
We use Nsight Systems for kernel time and Nanotron logs for step time, enabling end-to-end and micro-level attribution of speedups.

\begin{figure*}[t]
  \centering
  \includegraphics[width=\textwidth]{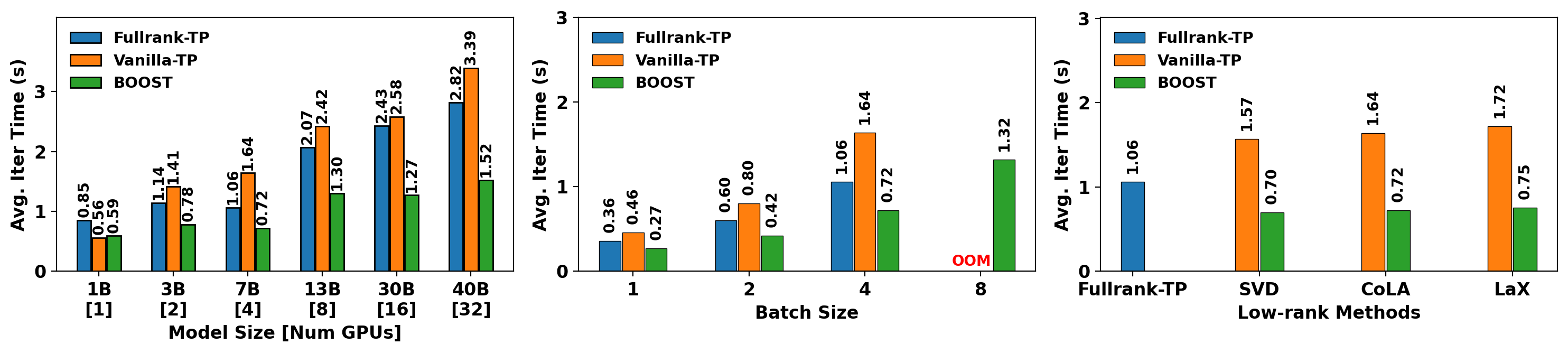}
  \vspace{-10pt}
  \caption{System-wide scalability and generality. (Left) Average iteration time on scaling model sizes with \#GPUs; (Middle) Average iteration time on scaling micro-batch size; (Right) Average iteration time on different low-rank architecture.}
  \label{fig:Iter-time}
\vspace{-10pt}
\end{figure*}

\section{Evaluation}
\subsection{Experimental Setup}
\label{sec:experimental-setup}
\textbf{Testbed.} 
Our evaluations are conducted on the NERSC-Perlmutter supercomputer. Each node contains 192 CPU cores, 256~GB DDR4 memory, and 4$\times$A100 GPUs, each containing 80GB HBM2 memory (320GB per node), with intra-node connection NVLink Gen3 and inter-node Slingshot~11. Following practice in BLOOM~\cite{workshop2022bloom}, OPT~\cite{zhang2022opt}, and LLaMA~\cite{dubey2024llama}, we use tensor-parallelism for scaling within a node, and pipeline-parallelism for scaling across nodes. Because our optimizations mainly target TP and apply per pipeline stage, we report results on up to 4 nodes (16 GPUs) to highlight per-node performance and resource utilization use.

\subsection{Methodology}

\textbf{Compared Approaches.} Throughout our experiments, we compare three different tensor-parallellism strategies. \fullrank is representative of real-world TP deployments that do not perform low-rank compression, due to which they incur higher memory and runtime overheads. \vanilla is a direct adaptation of the tensor-parallelism strategy proposed in Megatron-LM~\cite{shoeybi2019megatron} for low-rank architectures such as SVD, CoLA~\cite{liu2025cola}, and LAX~\cite{zhang2025lax}, which show reduced memory consumption and faster runtimes compared to full-rank in single GPU case. 
Lastly, we compare our proposed approach-- \framework, which incorporates the design principles discussed in \S~\ref{sec:design} to accelerate the training of low-rank bottleneck LLM architectures using tensor-parallelism.


\textbf{Models, Dataset, and Runtime Configuration.} 
We evaluate 6 different models of sizes 1B, 3B, 7B, 13B, 30B and 40B parameters, all from the LLaMA-2 family under pure bf16 training.
For system performance checks we use WikiText dataset with sequence length 4096, matching real-world settings~\cite{lang2024endtoend}.
Unless stated, we use a batch size of 4, tensor-parallelism of degree 4 (equal to the number of GPUs per node), pipeline and data-parallelism of 1, and disable activation checkpointing for faster iterations.
As a representative of bottleneck architecture LLM, we primarily use CoLA~\cite{liu2025cola} as its algorithmic effectiveness has been most thoroughly validated across different bottleneck architectures. Nonetheless, we demonstrate the generalizability of our proposed design principles to other low-rank approaches such as SVD~\cite{LORO_iclr2025,wang2025svdllm}, and LAX~\cite{zhang2025lax}.
For each experiment, we run 10 training iterations, of which the first 2 are considered as warm-up, and we report an average of the remainder 8 iterations.


\textbf{Key Performance Metrics.} 
Throughout our evaluations, we study the compared approaches using different metrics. First, we consider the \emph{Average Iteration Time}, which captures the holistic performance impacts, and is representative of the end-to-end time taken throughout the training process. Next, we measure the \emph{Kernel Execution Time} to break down the computational time taken by different GEMM operations, which also yields the corresponding FLOPS for understanding the kernel efficiency. Third, we study the \emph{Hardware Utilization}, measured as a percentage of the peak computational capability of the GPU, to understand resource usage by different approaches. Fourth, to study TP communication overheads, we measure the \emph{communication volume and time} of data transfers. Finally, we measure the \emph{memory saving and runtime overheads} when enabling activation checkpointing.

\subsection{Overall Training Performance}
\textbf{Scaling Model Sizes.}
In our first set of experiments, we evaluate the impact of weak scaling model sizes from 1B to 30B. For smaller models-- 1B, 3B, 7B, that fit in a single node, we keep increasing the tensor-parallelism degree, and beyond the 7B model, we use pipeline parallelism of degree 2, 4 and 8 for the 13B, 30B and 40B models, respectively. We use the iteration time to study the scalability behavior for different model sizes.

As shown in Figure~\ref{fig:Iter-time}(left), the 1B model (with no TP) is $1.4\times$ faster with low rank, confirming baseline efficiency of bottleneck architectures for LLMs. On scaling
to 3B model with 2 GPUs or higher, \fullrank\ outperforms \vanilla\ by $\sim$25\% because vanilla’s extra collectives and low-A.I.\ kernels offsets FLOP savings (§\ref{sec:bottleneck-tp}). In contrast, \framework\ achieves up to $1.91\times$ over \fullrank\ and $2.28\times$ over \vanilla. Even with cross-node PP at 13B/30B/40B, \framework’s TP optimizations significantly reduce iteration time, demonstrating end-to-end scalability.



\textbf{Scaling Batch Sizes.}
In the next set of experiments, we study the impact of increasing batch sizes on different approaches. We consider the 7B model running with a tensor-parallelism degree of 4, and scale the micro-batch size from 1$\cdots$8, until every approach runs out of memory (OOM). As depicted in Figure~\ref{fig:Iter-time}(middle), with increasing batch sizes of 1, 2, and 4, we see \framework outperforms the \fullrank by 1.3$\times$, 1.42$\times$, and 1.48$\times$, respectively, showing a linear acceleration in performance due to better hardware utilization of larger batches. Similar to scaling of models, \vanilla shows the worst performance across the compared approaches because it collects full hidden states and holds redundant activations (\S~\ref{sec:bottleneck-tp}), while \framework avoids these memory overheads, preserving the memory efficiency of low-rank bottleneck architectures for tensor-parallel setups, resulting in accommodating larger batch sizes. Finally, we observe that \framework's reduced memory footprint allows us to train larger batches while showing a superlinear scalability trend.

\textbf{Extending Across Different Bottleneck Architectures.}
Next, we demonstrate the generality of our approach across different LLM bottleneck architectures- SVD, CoLA, and LaX. We use the 7B model with 4 GPUs and compare the average iteration time for all approaches. Given the diversity of factorization methods used for producing low-rank architectures, through this experiment, we study whether the different underlying bottleneck structure of transformer blocks impact training performance when scaling across tensor-parallel ranks. As seen in Figure~\ref{fig:Iter-time}(right), \framework outperforms \vanilla and \fullrank approaches by 1.5$\times$ and 2.2$\times$, respectively, for all low-rank approaches. The slight variations in the iteration times are a result of differences in their respective low-rank architectures-- SVD runs fastest because of no intervening operations, CoLA is slower due to a nonlinearity between the low-rank linear layers, and LaX is slowest owing to an added residual path.
These results underpin the versatility of the design formulations proposed in \fullrank to optimize the communication and computations across different bottleneck architectures, irrespective of how the layers were factorized.

\begin{figure*}[t]
  \centering
  \includegraphics[width=\textwidth]{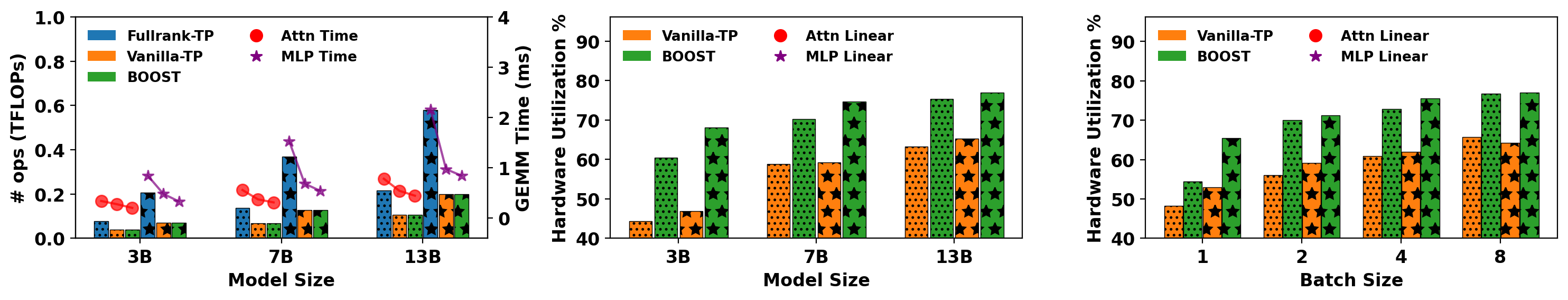}
  \vspace{-10pt}
  \caption{Computation Efficiency. (Left) Linear layer FLOPs and GEMM kernel time under different TP designs; (Middle) Hardware utilization of vanilla TP and BOOST for each linear layer; (Right) Hardware utilization on scaling micro-batch size for each linear layer.}
  \label{fig:computation-efficiency}
\end{figure*}

\begin{figure*}[t]
  \centering
  \includegraphics[width=\textwidth]{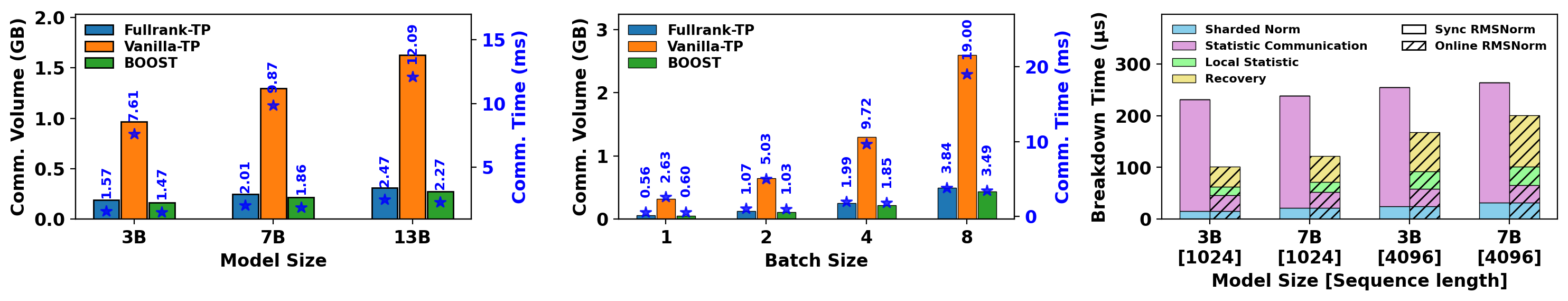}
  \vspace{-10pt}
  \caption{(Left) Communication volume and time on different TP strategy; (Middle) Communication volume and time on scaling micro-batch size; (Right) Online-RMSNorm breakdown.}
  \label{fig:communication-efficiency}
  \vspace{-10pt}
\end{figure*}

\subsection{Computation Efficiency}
In the next series of experiments, we study the computation efficiency introduced by \framework and quantify its impact on GEMM time and hardware utilization. We instrument each linear layer (attention QKV/out and MLP gate/up/down) in LLaMA-7B/13B and compare \fullrank, \vanilla, and our approach \framework. For each linear layer, we report (i) theoretical work (FLOPs) alongside measured GEMM time, and (ii) the resulting hardware utilization. This analysis explains why our design accelerates computation and verifies alignment with the predictions in \S~\ref{sec:bottleneck-tp}.

Figure~\ref{fig:computation-efficiency}~(left) shows that both low-rank designs, including \vanilla and \framework, perform substantially fewer FLOPs and achieve shorter GEMM times than \fullrank. The reductions are more pronounced in the MLP linears, where the factorization targets the larger $d_{ff}$ dimension, yielding a bigger drop in work than in attention. This explains why low-rank methods are faster than fullrank in computation. Importantly, \vanilla and \framework have the same FLOP counts and do the same amount of math.

Figure~\ref{fig:computation-efficiency} (middle) shows why \framework finishes the GEMMs faster than \vanilla despite their identical FLOP counts. From the figure we can see that \framework sustains higher hardware utilization in both attention and MLP linears. The improvement comes from a more hardware-friendly shard layout: \framework splits along the larger dimension, yielding GEMMs with larger reduction dimension, which are more compute-bound and less memory-bound. This aligns well with our arithmetic intensity analysis in \S~\ref{sec:bottleneck-tp}.

From Figure~\ref{fig:computation-efficiency} (right), we observe that as the micro-batch size increases, utilization rises for all methods because of more work per launch and better arithmetic intensity. \framework remains consistently higher than \vanilla across batches and layers, demonstrating the efficiency of its kernels to stay closer to the compute-bound regime and achieving end-to-end speedups.

Together, these results show that BOOST not only preserves the algorithmic efficiency of low-rank models, but also substantially improves hardware utilization—bridging the gap between theoretical compression and practical speedup.

\subsection{Communication Efficiency}

We instrument TP collectives and report the total communication volume per decoder block per pass and the corresponding wall-clock communication time. We compare \fullrank, \vanilla, and \framework across 7B/13B model sizes and varying micro-batch sizes.

From Figure~\ref{fig:communication-efficiency} (left), we can see that \vanilla communicates far more than \fullrank because each bottleneck block adds an extra TP chunk, making two activation all-reduces per block with payloads at the full hidden size $d$ or even the intermediate size $d_{ff}$ (\S~\ref{sec:bottleneck-tp}). This inflates both communication volume and time. In contrast, BOOST also increases the number of collectives, but performs them on low-rank activations ($bsr$ instead of $bsd$), leading to a communication volume smaller than \fullrank and orders of magnitude below \vanilla. The measured time reflects the same reduction, up to 8$\%$ faster than \fullrank and 5.3$\times$ faster than \vanilla in communication.

As the micro-batch size grows (Figure~\ref{fig:communication-efficiency} (middle)), communication volume increases roughly linearly for all methods. The slope for \vanilla is steep because of extra full-hidden activations. \framework’s growth grows much gentler because it communicates on low-rank payloads, and it remains consistently faster than \fullrank across batch sizes.

These measurements show that our \framework substantially reduces TP communication volume and time compared to \vanilla, and keeps it below \fullrank. This communication efficiency is a key contributor to the end-to-end throughput gains observed as models and batch size scale.


\subsection{Optimization Ablations}
\label{sec:optimization_ablations}
\paragraph{Online RMSNorm.}
Repositioning the TP chunk places sharded-unsafe RMSNorm inside the chunk, we therefore compare \emph{Sync} vs.\ \emph{Online} RMSNorm. Communication overhead for Online RMSNorm is measured as the difference between communication time with fusion and without fusion; per-step breakdowns are in Figure~\ref{fig:communication-efficiency}(right).

From this figure, we can observe that: 1) Normalization compute on sharded activations is effectively identical for both methods, the performance gap stems from communication, not arithmetic. 2) Sync RMSNorm is dominated by statistic communication, although its volume is tiny, it suffers high latency from kernel-launch overhead and poor effective bandwidth, so the pink bar is large and nearly flat across model sizes. With longer sequences, it grows modestly for the same reason, it is now launch/latency-bound instead of throughput-bound. 3) Online RMSNorm introduces minimal communication overhead because the statistic exchange is fused into the TP collective, adding little extra traffic when bandwidth is already saturated. Its runtime increase with model size/sequence length is driven primarily by the extra compute in recovery step, not by communication.

Overall, this highlights a clear communication–computation tradeoff: Online RMSNorm replaces the latency-dominated statistic exchange with a fused path, dramatically reducing communication time; the modest extra compute is more than offset by the lower communication cost.


\paragraph{Linear Layers Grouping.}
\begin{table}[t]
\caption{Per-decoder-block time (µs) for Default vs.Grouped linear-layer on CoLA LLaMA-7B model.}
\vspace{5pt}
\centering
\scriptsize
\setlength{\tabcolsep}{5pt}
\begin{tabular}{l *{3}{c} *{3}{c}}
\toprule
& \multicolumn{3}{c}{\textbf{bz=1}}
& \multicolumn{3}{c}{\textbf{bz=4}} \\
\cmidrule(lr){2-4}\cmidrule(lr){5-7}
\textbf{Block / Kernel}
& \textbf{Non.} & \textbf{Grouped} & \textbf{Speed}
& \textbf{Non.} & \textbf{Grouped} & \textbf{Speed} \\
\midrule
MLP1 Comp  & 355   & 292   & 1.22$\times$      & 1115 & 1082 & 1.03$\times$ \\
MLP1 Comm  & 266   & 218   & 1.22$\times$      & 620 & 580 & 1.07$\times$ \\
QKV Comp   & 391  & 255  & 1.53$\times$   & 939 & 877 & 1.07$\times$ \\
QKV Comm   & 406   & 288   & 1.41$\times$      & 981 & 806 & 1.22$\times$ \\
\midrule
\textbf{Layerwise Total}
           & \textbf{2773}   & \textbf{2395}   & \textbf{1.16$\times$}      & \textbf{7577} & \textbf{7266} & \textbf{1.04$\times$} \\
\bottomrule
\end{tabular}
\label{tab:linear_grouping}
\vspace{-10pt}
\end{table}

As shown in Table~\ref{tab:linear_grouping}, grouping improves both compute and communication for the QKV projection and the MLP gate–up path. On compute, grouping amortizes activation reads and enlarges the effective GEMM, raising A.I. and utilization (e.g., 1.53× for QKV at bz=1, 1.07× at bz=4). On communication, it reduces kernel launches and increases per-call payload, improving effective bandwidth (e.g., 1.41× for QKV at bz=1, 1.22× at bz=4). Aggregated per block, this yields a 1.16× speedup at bz=1 and 1.04× at bz=4, confirming that grouping translates analytical gains into end-to-end savings across batch sizes. 

Note that the gains are larger at bz=1 than at bz=4. With a smaller per-GPU workload, kernels are more memory-bound and exhibit lower hardware utilization, so grouping’s effects in larger effective GEMMs and fewer launches will translate into a bigger speedup.

\paragraph{Efficient Activation Checkpointing.}
Table~\ref{tab:memory_breakdown} provides a comparative per-TP-rank breakdown of memory utilization across methods. The choice of TP strategy does not affect the persistent memory used by weights, gradients, or optimizer states. Instead, \vanilla\ incurs additional activation and communication-buffer memory due to redundant full-width activations materialized after the row-parallel all-reduce, together with larger buffers for full-rank collectives, whereas \framework\ operates at the smaller bottleneck dimension $r$. 

Table~\ref{tab:activation_checkpointing} compares low-rank activation checkpointing under \vanilla\ and \framework. For each batch size, we report activation memory saved by checkpointing ($\Delta\text{Mem}$), extra re-forward time ($+\text{Time}$), and efficiency $\mathrm{Eff}_{\text{ckpt}}=\Delta\text{Mem}/(+\text{Time})$. Larger values indicate more memory saved per unit of re-forward.

Our method consistently yields higher $\mathrm{Eff}_{\text{ckpt}}$: for bz=4, 193.5 MB/ms for \framework vs.\ 113.7~MB/ms for \vanilla (\textbf{1.70$\times$}); for bz=8, 177.0 for \framework vs.\ 113.6~MB/ms for \vanilla (\textbf{1.56$\times$}). Although the absolute $\Delta$Mem of the \framework is smaller because \vanilla incurs extra activation memory (\S~\ref{sec:bottleneck-tp}), its substantially lower re-forward cost ($+$Time) yields a higher $\mathrm{Eff_{ckpt}}$. 

This gain of $\mathrm{Eff_{ckpt}}$ stems from \framework eliminating extra communication in the re-forward path (§\ref{sec:activation checkpointing}), making activation checkpointing more efficient.




\begin{table}[t]
\caption{Per-TP-rank memory breakdown on CoLA LLaMA-7B (bz=4, seq\_len=4k). Memory is reported in GB.}
\vspace{5pt}
\centering
\small
\begin{tabular}{lccccc}
\toprule
\textbf{Method} & \textbf{Wgt.} & \textbf{Grad.} & \textbf{Opt.} & \shortstack{\textbf{Act.+ others}} & \textbf{Total} \\
\midrule
Vanilla TP & 1.25 & 1.25 & 2.50 & 30.73 & 35.73 \\
Ours & 1.25 & 1.25 & 2.50 & 22.14 & 27.14 \\
\bottomrule
\end{tabular}
\label{tab:memory_breakdown}
\vspace{-10pt}
\end{table}

\begin{table}[t]
\caption{Activation checkpointing efficiency on LLaMA-7B.
$\Delta$Mem and $+\,$Time are measured with no ckpt. at the same batch size.
$\mathrm{Eff_{ckpt}}=\Delta\text{Mem}/(+\text{Time})$ (MB per ms; higher is better).}
\vspace{5pt}
\centering
\resizebox{\linewidth}{!}{%
\begin{tabular}{l c c c c}
\toprule
\textbf{Method} & \textbf{bz} & $\boldsymbol{\Delta}$\textbf{Mem (MB)} & $\boldsymbol{+}$\textbf{Time (ms)} & $\boldsymbol{\mathrm{Eff_{ckpt}}}$ \textbf{(MB/ms)}\\
\midrule
Vanilla & 4 & 26{,}022 & 229  & 113.7\\
Ours     & 4 & 17{,}414 &  90  & \textbf{193.5}\\
\midrule
Vanilla & 8 & 52{,}280 & 460  & 113.6 \\
Ours    & 8 & 35{,}392 & 200  & \textbf{177.0}\\
\bottomrule
\end{tabular}
}
\label{tab:activation_checkpointing}
\end{table}


\section{Discussion and Future Work}
\label{sec:discussion}

In this work, we focus on dense low-rank bottleneck architectures, since current low-rank pre-training methods primarily target this regime. However, the core motivation behind BTP is broader. At a high level, BTP applies whenever a submodule becomes deeper, introducing more tensor-parallel synchronization points, and contains a narrow bottleneck activation, allowing TP collectives to be shifted to the lowest-dimensional intermediate representation.

More generally, BTP does not require strictly paired down/up projections or a fixed global bottleneck rank. As long as a module contains a well-defined narrow activation region, TP collectives can in principle be placed there, even in the presence of unpaired or non-uniform bottlenecks. This suggests that the framework may extend beyond the specific bottleneck layouts studied in this paper.

BTP can also be extended to architectures with grouped-query attention (GQA). In bottlenecked architectures, each linear transformation is replaced by a deeper sequence of low-rank linear layers. Under GQA, the required modification is simply to adjust the shapes of the Q, K, and V projections to respect the grouped-head layout. Importantly, these shape changes leave the core BTP mechanism and its communication-placement strategy unchanged.

Another promising direction is to extend BTP to mixture-of-experts (MoE) models. BTP is orthogonal to expert parallelism: EP routes tokens across experts, whereas BTP changes tensor-parallel boundary placement within a deeper, narrower bottlenecked submodule, provided the shifted TP chunk can still be executed correctly on sharded activations. In current MoE designs, experts are often made more fine-grained without introducing such bottlenecks, which can further increase token-routing all-to-all communication. However, if the experts themselves, or the dense/shared components, are bottlenecked or large enough to require tensor parallelism, BTP can still be applied to shift TP collectives to the narrow intermediate activation, thereby reducing TP communication while preserving higher arithmetic intensity. This is compatible with EP because token-routing all-to-all operates on sharded activations and is therefore orthogonal to TP boundary placement. Since TP remains common in MoE systems, often together with EP, understanding BTP--EP interactions under realistic workloads is an important direction for future work.



\section{Conclusion}
We have introduced \framework{}, a framework for efficiently training large-scale low-rank bottleneck transformer models. By analyzing the computation and communication characteristics of bottleneck architectures, we identified key inefficiencies in naive tensor parallelism, including low arithmetic intensity and excessive communication volume. Our Bottleneck-aware Tensor Parallelism (BTP) design addresses these challenges by redefining TP chunk boundaries to raise GEMM efficiency and cut communication, while Online RMSNorm, linear grouping, and low-rank checkpointing further reduce launches and syncs. Across multiple low-rank architectures and model sizes, \framework{} delivers consistent speedups over full-rank TP and vanilla low-rank TP, approaching high hardware utilization while preserving compression benefits, making it a practical, scalable solution for low-rank LLM pre-training.

\section*{Acknowledgements}
We thank the anonymous reviewers and our shepherd for their thoughtful and constructive feedback, which substantially improved this paper. We also thank Linxing Jiang for helpful discussions on the initial ideas and for insightful advice on framing of the paper.

This material is based upon work supported by the U.S. Department of Energy, Office of Science, Office of Advanced Scientific Computing Research, Artificial Intelligence for Science program, under contracts DE-SC0025390 and DE-AC02-06CH11357. This research used resources of the National Energy Research Scientific Computing Center, a DOE Office of Science User Facility supported by the Office of Science of the U.S. Department of Energy under Contract No. DE-AC02-05CH11231 using NERSC award ASCR-ERCAP0030039, as well as NERSC award ALCC-ERCAP0031379.





\bibliography{mlsys2026/main}
\bibliographystyle{mlsys2025}

\newpage
\appendix
\section{Artifact Appendix}

\subsection{Abstract}

This artifact contains the implementation of BOOST, the training framework proposed in the MLSys 2026 paper "BOOST: Bottleneck-Optimized Scalable Training Framework for Low-Rank Large Language Models." The artifact includes the source code, experiment scripts, and configuration files used to reproduce the key experimental results in the paper. 

The artifact implements the core components of BOOST, including Bottleneck-aware Tensor Parallelism (BTP) and Online RMSNorm integrated into a distributed training framework based on PyTorch and Nanotron. These components enable efficient distributed training of low-rank bottleneck architectures for large language models. 

The artifact supports the main claims of the paper regarding improved training efficiency and scalability for low-rank LLM architectures. In particular, it demonstrates reduced communication overhead, improved GPU utilization, and faster training iteration times compared with both full-rank tensor-parallel baselines and naive tensor-parallel implementations of low-rank models. Minimal hardware requirements are 1–4 NVIDIA GPUs with CUDA support and a Linux environment with PyTorch and NCCL installed. 

The artifact provides scripts to install dependencies, run small-scale distributed training experiments, and reproduce representative performance metrics such as iteration time, communication volume, and hardware flops utilization. Running the provided scripts generates logs and profiling outputs demonstrating the performance improvements of BOOST over baseline implementations.

\subsection{Artifact check-list}


{\small
\begin{itemize}
  \item {\bf Algorithm: Distributed training for low-rank LLMs; bottleneck-aware tensor parallelism}
  \item {\bf Dataset: WikiText / synthetic short-run data for validation}
  \item {\bf Run-time environment: Nanotron; Linux; CUDA; multi-GPU distributed training}
  \item {\bf Hardware: 80GB A100 NVIDIA GPUs, 4 GPUs per node}
  \item {\bf Execution: shell scripts; torchrun}
  \item {\bf Metrics: iteration time; communication volume; hardware utilization; throughput}
  \item {\bf Output: training logs; summary tables}
  \item {\bf How much time is needed to prepare workflow (approximately)?: 10 minutes}
  \item {\bf How much time is needed to complete experiments (approximately)?: 1 hour}
  \item {\bf Publicly available: Yes.}
  \item {\bf Code licenses: Apache-2.0}
  \item {\bf GitHub link: \url{https://github.com/Arcana-2236/BOOST}}
\end{itemize}

\subsection{Description}

\subsubsection{How delivered}
The artifact is delivered as a public GitHub repository containing the BOOST implementation, example configurations, experiment scripts, and instructions for installation and evaluation. The repository includes:
\begin{itemize}
    \item the BOOST implementation integrated into the Nanotron training framework,
    \item configuration files for full-rank, vanilla low-rank, and BOOST runs,
    \item scripts for lightweight validation experiments,
\end{itemize}
The public repository URL is provided in the submission form and in the README.

\subsubsection{Hardware dependencies}
The minimal supported setup is 1--4 NVIDIA GPUs with CUDA support. A single-node 4-GPU setup is sufficient for the main lightweight validation experiments. The main experiments in the paper were conducted on NVIDIA A100 GPUs (80GB) and larger multi-node settings, but the artifact includes smaller validation runs intended for AE reviewers.

Recommended:
\begin{itemize}
    \item 4 NVIDIA A100 GPUs for the closest reproduction of the reported results;
    \item NVLink-enabled intra-node communication for tensor-parallel evaluation.
\end{itemize}

\subsubsection{Software dependencies}
The artifact requires:
\begin{itemize}
    \item Linux,
    \item Python 3.10 or newer,
    \item CUDA 12.x,
    \item PyTorch with CUDA support,
    \item NCCL,
    \item Nanotron.
\end{itemize}

Optional dependencies:
\begin{itemize}
    \item Nsight Systems for profiling,
    \item Conda or docker for environment management.
\end{itemize}

\subsection{Installation}

The artifact can be executed using the NVIDIA PyTorch container to ensure
consistent dependencies across environments. Detailed installation
instructions are provided in the GitHub repository README.

\subsection{Experiment workflow}

The repository provides scripts for running distributed training
experiments comparing three configurations: the full-rank baseline,
the low-rank vanilla tensor-parallel baseline, and the BOOST approach
with Bottleneck-aware Tensor Parallelism (BTP).

To reproduce a direct comparison between the three approaches, the
repository provides a script that sequentially runs the full-rank
baseline, the low-rank vanilla tensor-parallel baseline, and the BTP
implementation. The script records the average iteration time for each
configuration, enabling a quick comparison of training performance.

The comparison experiment can be launched using the following command:
\begin{itemize}
\item bash run\_iter\_compare.sh
\end{itemize}

Alternatively, the configurations can be executed individually using
the following commands:

\begin{itemize}

\item Full-rank baseline

CUDA\_DEVICE\_MAX\_CONNECTIONS=1 torchrun --nproc\_per\_node=4 run\_train.py \\
--config-file examples/config\_tiny\_llama.yaml

\item CoLA model with BOOST (BTP)

CUDA\_DEVICE\_MAX\_CONNECTIONS=1 torchrun --nproc\_per\_node=4 \\
examples/cola/train\_cola.py \\
--config-file examples/cola/config\_tiny\_cola\_llama.yaml

\end{itemize}

\subsection{Evaluation and expected result}

After running the experiment workflow, logs and runtime statistics are
generated in the logging directory.

Reviewers should observe the qualitative trends reported in the paper,
including:

\begin{itemize}

\item BOOST achieves lower iteration time than naive tensor-parallel
low-rank implementations.

\item Communication overhead is reduced due to communication on
low-rank activations.

\item Hardware utilization improves compared with naive tensor-parallel
baselines.

\end{itemize}

Exact numerical values may vary depending on GPU model and system
configuration, but the relative performance trends should remain
consistent with those reported in the paper.

The experiments reported in the paper were conducted on 80GB NVIDIA A100 GPUs. 
For reproduction on 40GB A100 GPUs, the batch size can be adjusted accordingly 
using the provided scripts.

\subsection{Experiment customization}

The artifact provides .yaml configuration files under the
\texttt{examples/} directory that allow users to modify
experimental parameters.

Users may customize:

\begin{itemize}

\item model size (TinyLLaMA, LLaMA-1B, LLaMA-7B, etc.)

\item batch size and sequence length

\item 3D parallelism configurations

\item low-rank model variants and BOOST optimizations

\end{itemize}

These parameters can be modified by editing the configuration files
before launching training runs. Additional instructions for running
larger experiments and profiling workflows are provided in the GitHub
README.

\subsection{Methodology}

Submission, reviewing, and badging methodology:

\begin{itemize}
\item \url{http://cTuning.org/ae/submission-20190109.html}
\item \url{http://cTuning.org/ae/reviewing-20190109.html}
\item \url{https://www.acm.org/publications/policies/artifact-review-badging}
\end{itemize}

\newpage

\begin{table*}[t]
\caption{
Per iteration communication volume ($V_{\text{comm}}$) under different parallelism strategies for full-rank and bottleneck architecture in Llama-like models.
}
\label{tab:comm-volume-table}
\vskip 0.15in
\begin{center}
\begin{small}
\begin{sc}
\begin{tabular}{lccc}
\toprule
Strategy & Full-Rank & \multicolumn{2}{c}{Bottleneck Architecture} \\
         &           & Vanilla & Ours \\
\midrule
Data Parallel (DP)     & $l(4d^2+3dd_{ff})$ & \multicolumn{2}{c}{$l(11dr+3d_{ff}r)$} \\
Pipeline Parallel (PP) & $2pbsd$ & \multicolumn{2}{c}{$2pbsd$} \\
Tensor Parallel (TP)   & $2l(2bsd)$ & $2l(5bsd+2bsd_{ff})$ & $2l(7bsr)$ \\
\bottomrule
\end{tabular}
\end{sc}
\end{small}
\end{center}
\vskip -0.1in
\end{table*}

\begin{table*}[t]
\caption{
Per MLP block GEMM arithmetic intensity (A.I.) for fullrank TP, vanilla low-rank TP and bottleneck-aware TP (Assuming $d_{ff}=\alpha d$, $d=\beta r$).
}
\label{tab:arithmetic-intensity-analysis}
\vskip 0.15in
\renewcommand{\arraystretch}{1.8} 
\begin{center}
\begin{small}
\begin{sc}
\begin{tabular}{lccc}
\toprule
TP Design & FLOPs & Data Movement & A.I. \\
\midrule
Fullrank TP & $4\alpha bsd^2/TP$ &  $4d(bs+\frac{\alpha(d+bs)}{TP})$ & $\frac{\alpha bsd^2}{bsdTP+\alpha d(d+bs)}$\\
Vanilla Low-rank TP & \multirow{2}{*}{$\frac{4(1+\alpha)bsd^2}{\beta TP}$} & $4d((1+\alpha)bs+\frac{(1+\alpha)d+2bs}{\beta TP})$ & $
\frac{4 b s d^2 (1 + \alpha)}{4 b s d \, \beta TP (1 + \alpha) + 4 d^2 (1 + \alpha) + 8 b s d}
$
\\
Bottleneck TP &  & $4d(\frac{(1+\alpha)(\beta bs+d)+2bsTP}{\beta TP})$ & $
\frac{4 b s d^2 (1 + \alpha)}{4 \beta b s d (1 + \alpha) + 4 d^2 (1 + \alpha) + 8 b s d \, TP}
$
 \\
\bottomrule
\end{tabular}
\end{sc}
\end{small}
\end{center}

\vskip -0.1in
\end{table*}

\section{Appendix}
\subsection{Detailed Performance Analysis}
\textbf{Communication volume analysis.}
We report per-iteration communication volume $V_{\text{comm}}$ for a LLaMA-style decoder in Table~\ref{tab:comm-volume-table} with $l$ layers, microbatch size $b$, sequence length $s$, hidden size $d$, MLP expansion $d_{ff}$, low-rank size $r\!\ll\!d$, and pipeline degree p. The counts assume Megatron-style implementations, no sequence parallelism, constants and embedding/norm traffic are omitted for clarity.

\emph{DP.} Gradient synchronization volume scales with parameter size. For full rank, each layer’s attention contributes $4d^2$ (Q,K,V,O) and MLP contributes $3dd_{ff}$ (gate, up, down), giving $l(4d^2+3dd_{ff})$. For low-rank variants (both Vanilla and Ours), replace $d^2\!\to\!2dr$ and $dd_{ff}\!\to\!dr+d_{ff}r$ consistent with factorized weights across the same 4+3 linear maps, yielding $l(11dr+3d_{ff}r)$.

\emph{PP.} Pipeline activations of size $[b,s,d]$ traverse stage boundaries in forward/backward. Abstracting per-edge factors and schedule details, volume scales as $pbsd$ (with $p$ stages) per pass, the same for full rank and low rank because boundary tensors are at width $d$ in this configuration.

\emph{TP.} In Megatron’s column–row pattern, each decoder block issues one activation-sized collective in attention and one in MLP per pass, i.e., $2\times[b,s,d]$ per layer. Accounting for forward and backward gives $2l(2bsd)$. Vanilla Low-rank TP treats each low-rank linear separately increases collective count. Per pass, attention triggers $\;4bsd\;$ (one per $d{\times}r$/$r{\times}d$ linear) and MLP triggers $\;bsd+2bs\,d_{ff}\;$ (down plus up), totaling $(5bsd+2bsd_{ff})$; forward+backward yields $2l(5bsd+2bsd_{ff})$. With BOOST, the TP chunk boundary is shifted so the single collective happens at low-rank tensor $[b,s,r]$. Summing over chunks gives $7\,bsr$ per pass and thus $2l(7bsr)$ per iteration. This replaces $O(d)$ activation-sized traffic with $O(r)$, explaining the large reduction relative to full rank and Vanilla low rank.

\textbf{Arithmetic Intensity analysis.}
We report per-MLP–block arithmetic intensity (A.I. = FLOPs / data moved) under three TP designs (Table~\ref{tab:arithmetic-intensity-analysis}. Let hidden size be $d$, expansion $d_{ff}=\alpha d$, low-rank size $r$ with $d=\beta r$ ($\beta\!=\!d/r$), microbatch $b$, sequence length $s$, and TP degree $\mathrm{TP}$. We count forward only GEMMs and the dominant activation/weight traffic; constants, bias/elementwise ops, and embedding/norm terms are omitted.

The result separates \emph{algorithmic} vs.\ \emph{system} effects. First, both low-rank variants (Vanilla and \framework{}) have strictly fewer FLOPs than Full-rank TP, this is the \textbf{algorithmic efficiency} of bottleneck factorization (\(d\times d \to d\times r,\, r\times d\) with \(r\!\ll\!d\)). Second, \emph{within} the low-rank family, \textbf{Vanilla and \framework{} have the same FLOPs}: the TP design does not change the amount of math, only how it is scheduled and communicated. Third, \framework{}’s advantage comes from \textbf{lower data movement}: by sharding along the large dimension and communicating on the low-rank path, \framework{} reduces activation traffic and yields higher arithmetic intensity (more FLOPs per byte). Hence, the performance gap between Vanilla and \framework{} is a \textbf{system-level} gain (communication/memory traffic), not an algorithmic one.

\subsection{Model Configuration}
Table~\ref{tab:model-config} lists the LLaMA-style model settings used in our experiments. For each size (1B–30B), we specify the number of transformer layers, hidden width $d$, and MLP expansion $d_{ff}$, alongside a canonical low rank $r=d/4$ for the bottleneck variants (SVD/CoLA/LaX). These configurations follow common scaling rules while keeping $r$ proportional to $d$ so that accuracy remains comparable and the system effects of low-rank factorization (communication volume and arithmetic intensity) are isolated. We use these fixed shapes across all TP/PP settings to ensure that reported performance differences stem from the parallelism design rather than architectural changes.

\begin{table}[t]
\caption{Model configuration (LLaMA-style). We use a canonical low rank $r\!=\!d/4$.}
\centering
\small
\setlength{\tabcolsep}{6pt}
\renewcommand{\arraystretch}{1.15}
\begin{tabular}{lrrrrr}
\toprule
\textbf{Model size} & \textbf{Layers} & \textbf{head} & $\boldsymbol{d}$ & $\boldsymbol{d_{ff}}$ & $\boldsymbol{r}$ \\
\midrule
1B   & 24 & 32 & 2048 & 5472 & 512 \\
3B   & 28 & 24 & 3072 & 8192 & 768 \\
7B   & 32 & 32 & 4096 & 11008 & 1024 \\
13B  & 40 & 40 & 5120 & 13824 & 1280 \\
30B  & 36 & 64 & 8192 & 22016 & 2048 \\
\bottomrule
\end{tabular}
\label{tab:model-config}
\end{table}

\subsection{Low-rank Training Algorithm}
\label{app:lowrank_training}

\paragraph{Common setup.}
Unless noted otherwise, we use pure \textbf{bf16} for training with \textbf{AdamW} and a single \textbf{fused} CUDA kernel that updates parameters and optimizer states. 

\paragraph{SVD baseline (system baseline).}
To establish a system-performance lower bound with minimal algorithmic changes, we \emph{mechanically} replace every full-rank linear $W$ with two low-rank linears:
\begin{equation}
y \;=\; A\,(B\,x), \quad A\in\mathbb{R}^{d_{\text{out}}\times r},\; B\in\mathbb{R}^{r\times d_{\text{in}}}.
\end{equation}
No nonlinearity is introduced between $B$ and $A$. This variant isolates the effect of reduced arithmetic/memory while keeping the computation graph closest to the full-rank case.

\paragraph{CoLA variant (nonlinear bottleneck).}
For CoLA, we insert a nonlinearity in the low-rank pathway. Concretely, with \emph{SwiGLU} as the nonlinearity, following the original paper setting,
\begin{equation}
y = A \cdot \operatorname{SwiGLU}(Bx).
\end{equation}

\paragraph{LaX variant (residual low-rank).}
For LaX, we implement a residual low-rank path using an \emph{Identity Gate} and inter-layer LaX (low-rank residual):
\begin{equation}
\begin{aligned}
h_{i-1} &= A_{i-1}\,x_{i-1} \in \mathbb{R}^{r},\\
h_i     &= A_i\,x_i \in \mathbb{R}^{r},\\
\tilde{y}_i &= B_i\,(h_i + h_{i-1}) \in \mathbb{R}^{d_{\mathrm{out}}}.
\end{aligned}
\end{equation}

\subsection{Linear Grouping Details}
\label{app:linear_grouping}
Figure~\ref{fig:layer_grouping} details our linear-grouping implementation. In BTP, the first down-projection is row-parallel; we therefore concatenate the weights along the rank dimension and execute a single GEMM to produce a wider activation, which is then split per path. The subsequent all-reduce remains unchanged. For the column-parallel up-projections, we stack weights/activations and use a batched GEMM (e.g., \texttt{einsum}) to accommodate the stacked shapes.

\begin{figure}[t]
 \centering
 \includegraphics[width=0.98\linewidth]{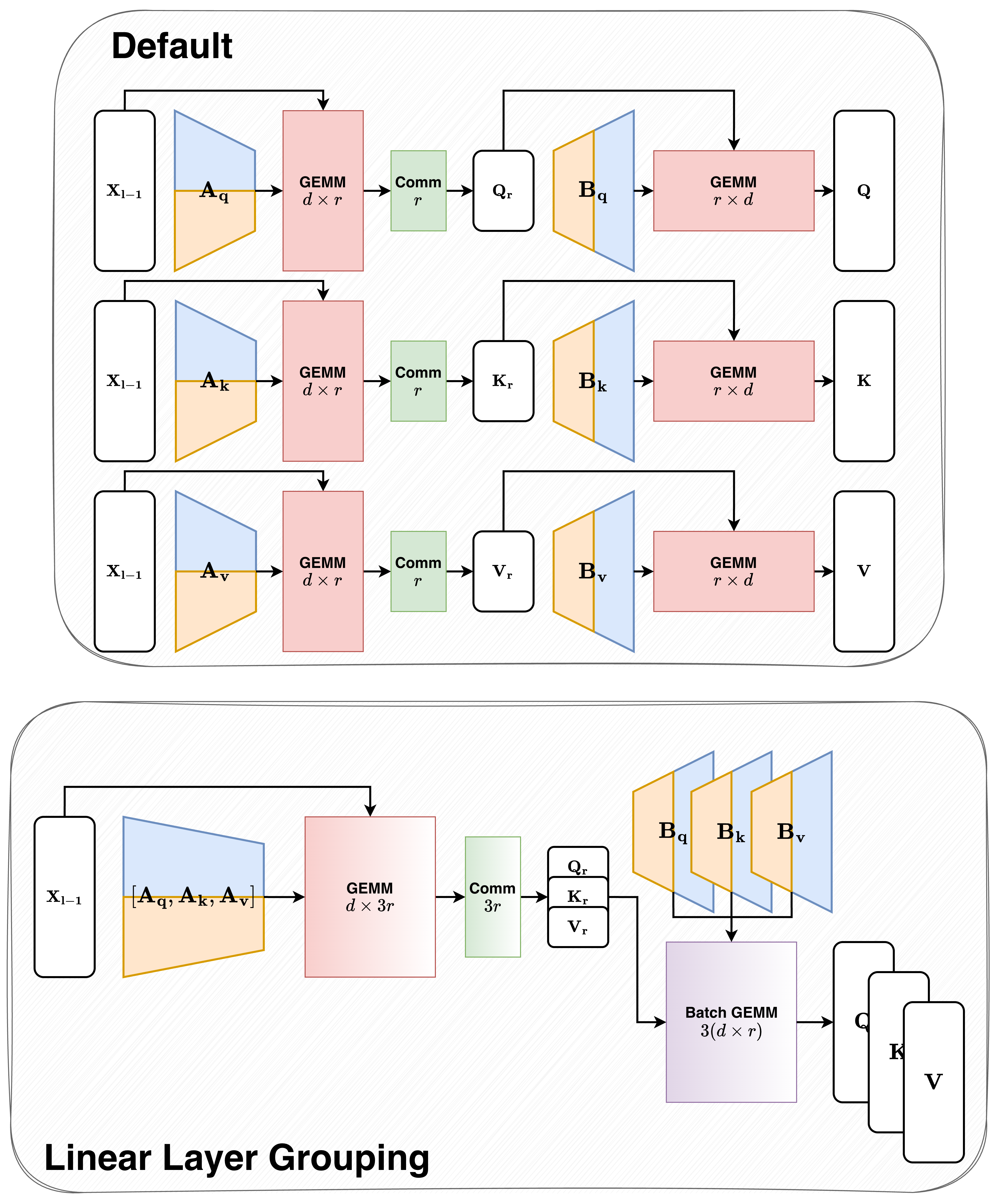}
 \caption{Linear layer grouping implementation}
 \label{fig:layer_grouping}
 \vspace{-10pt}
\end{figure}

\subsection{Framework Details}
\label{app:nanotron-overview}

\paragraph{Process groups and topology.}
Nanotron exposes a \textit{ParallelContext} that constructs and tracks all process groups for 3D parallelism (DP/TP/PP) alongside a world group. It provides rank mapping utilities (world\_rank\_matrix, local rank queries) and lets modules invoke collectives on the correct subgroup granularity.

\paragraph{Tensor parallelism (TP).}
Nanotron implements Megatron-style column/row-parallel linears, parallel vocabulary, cross-entropy over sharded logits, and distributed samplers. Different sharding patterns trade duplicated compute for reduced communication; engines select the appropriate collective (e.g., all-reduce or all-gather) over the TP group.

\paragraph{Pipeline parallelism (PP).}
Models are assembled from \emph{PipelineBlocks}, a lightweight wrapper that pins a submodule to a PP rank and routes tensors via \texttt{TensorPointer} when a rank is not responsible for compute. We use \emph{One-Forward-One-Backward (1F1B)} scheduling.

\paragraph{Parameter metadata.}
To make sharding and tying explicit, Nanotron augments \texttt{nn.Parameter} with metadata:
(i) \emph{Sharded parameters} carry \texttt{ShardedInfo} (global ranks, local/global slices, and the unsharded shape), enabling precise slicing and regrouping;  
(ii) \emph{Tied parameters} carry \texttt{TiedInfo} (a canonical name, scope root, involved ranks, and an optional \texttt{reduce\_op} for gradient syncing). A parameter may be both sharded (across TP) and tied (across PP endpoints).

\paragraph{Recomputation utilities.}
A \texttt{@checkpoint\_method} decorator enables activation recomputation on a per-module basis, reducing activation memory at the cost of controlled recompute. We implement the low-rank activation checkpointing with this.

\paragraph{On-device initialization.}
Context managers allow constructing modules directly on \texttt{cuda} (bf16) or on \texttt{meta} and then materializing on device, avoiding large CPU-side allocations and speeding up initialization.

\paragraph{Collectives.}
All DP/TP/PP communication uses NCCL collectives (e.g., all-reduce, all-gather, reduce-scatter) on a dedicated communication stream. Fine-grained overlap of all-reduce with compute is not currently supported in Nanotron.



\end{document}